\documentclass[10pt,twocolumn,letterpaper]{article}

\usepackage[pagenumbers]{wacv}      % To produce the REVIEW version for the algorithms track
\usepackage{graphicx}
\usepackage{amsmath}
\usepackage{amssymb}
\usepackage{booktabs}
\usepackage{booktabs}
\usepackage{array}
\usepackage{tabularx}

\usepackage[pagebackref,breaklinks,colorlinks]{hyperref}

\usepackage[capitalize]{cleveref}
\crefname{section}{Sec.}{Secs.}
\Crefname{section}{Section}{Sections}
\Crefname{table}{Table}{Tables}
\crefname{table}{Tab.}{Tabs.}

\def\wacvPaperID{6} % *** Enter the WACV Paper ID here
\def\confName{WACV}
\def\confYear{2025}

\title{Can Pose Transfer Models Generate Realistic Human Motion?}

\author{
Vaclav Knapp\\
SSPS \\
Prague, Czech Republic \\
{\tt\small Knapp.Va.2022@skola.ssps.cz}
\and
Matyas Bohacek \\
Stanford University \\
Stanford, CA, USA \\
{\tt\small maty@stanford.edu}
}

\begin{document}
\twocolumn[{

    \renewcommand\twocolumn[1][]{#1}%
    \maketitle

    \centering
    \begin{tabular}{c@{\hskip 4pt}c@{\hskip 4pt}c@{\hskip 25pt}c@{\hskip 4pt}c@{\hskip 4pt}c}

        \multicolumn{3}{c}{\small Consistent and photorealistic frames \vspace{0.1cm}} \hspace{26pt} &
        \multicolumn{3}{c}{\small Inconsistent or non-photorealistic frames} \\
        
        \includegraphics[width=0.15\textwidth]{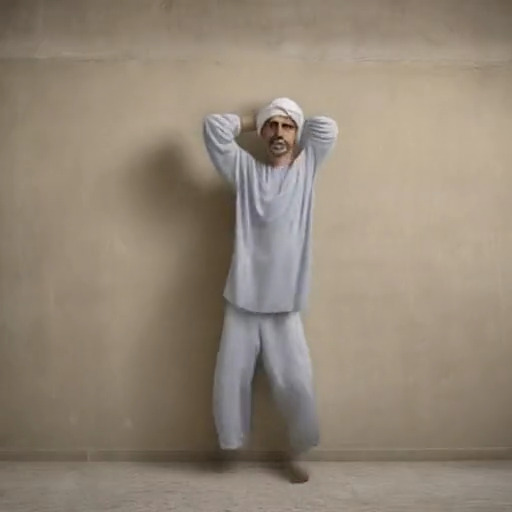} &
        \includegraphics[width=0.15\textwidth]{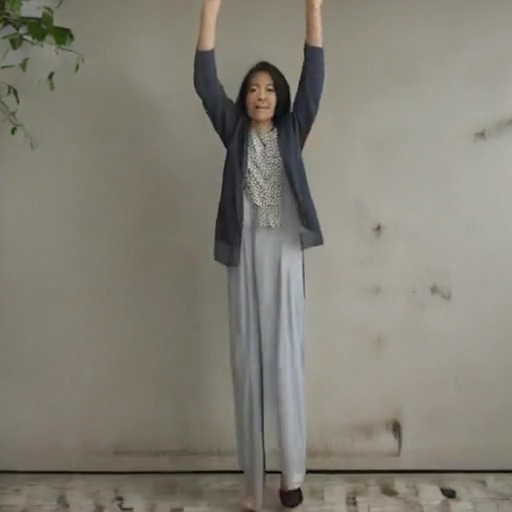} &
        \fbox{\includegraphics[width=0.15\textwidth]{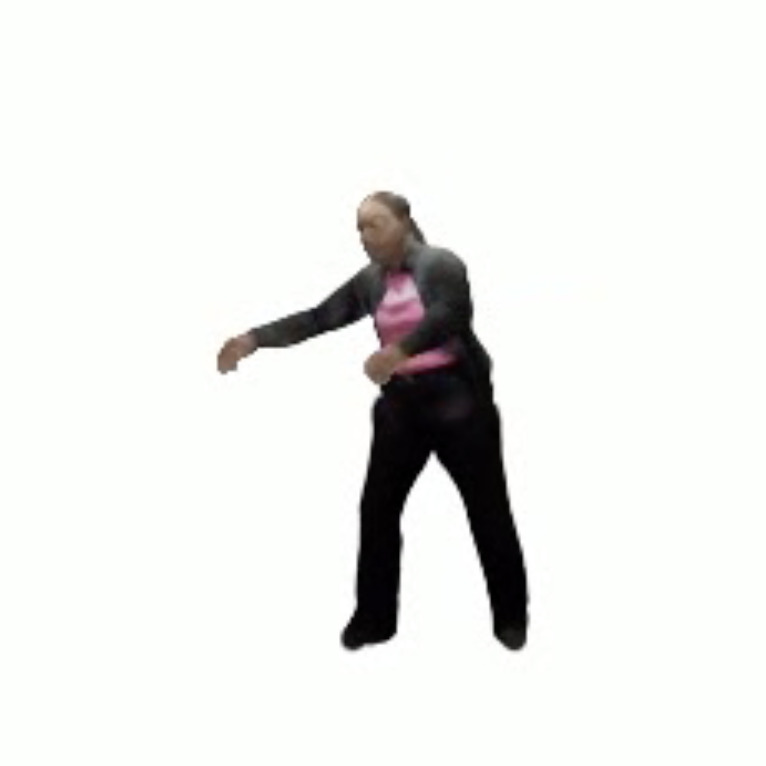}} &
        \includegraphics[width=0.15\textwidth]{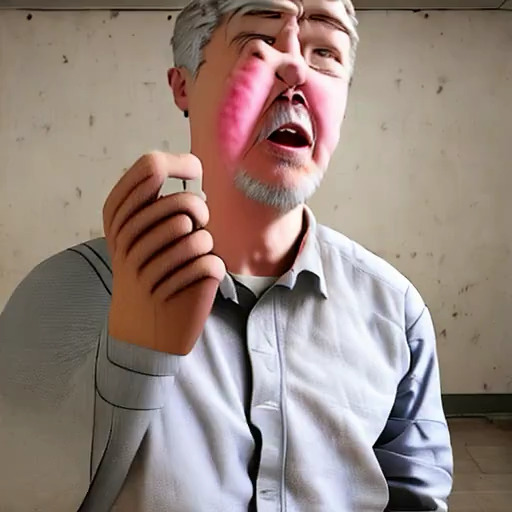} &
        \includegraphics[width=0.15\textwidth]{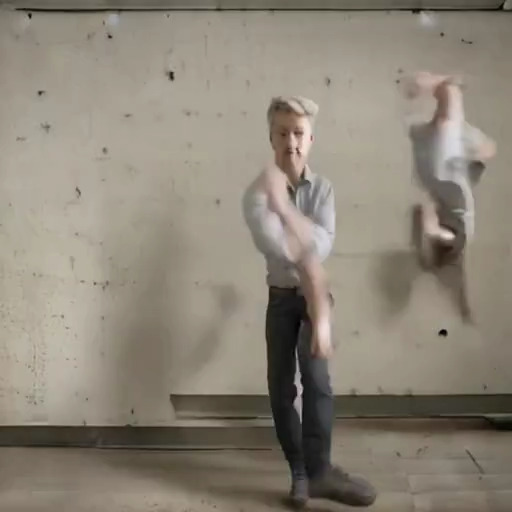} &
        \fbox{\includegraphics[width=0.15\textwidth]{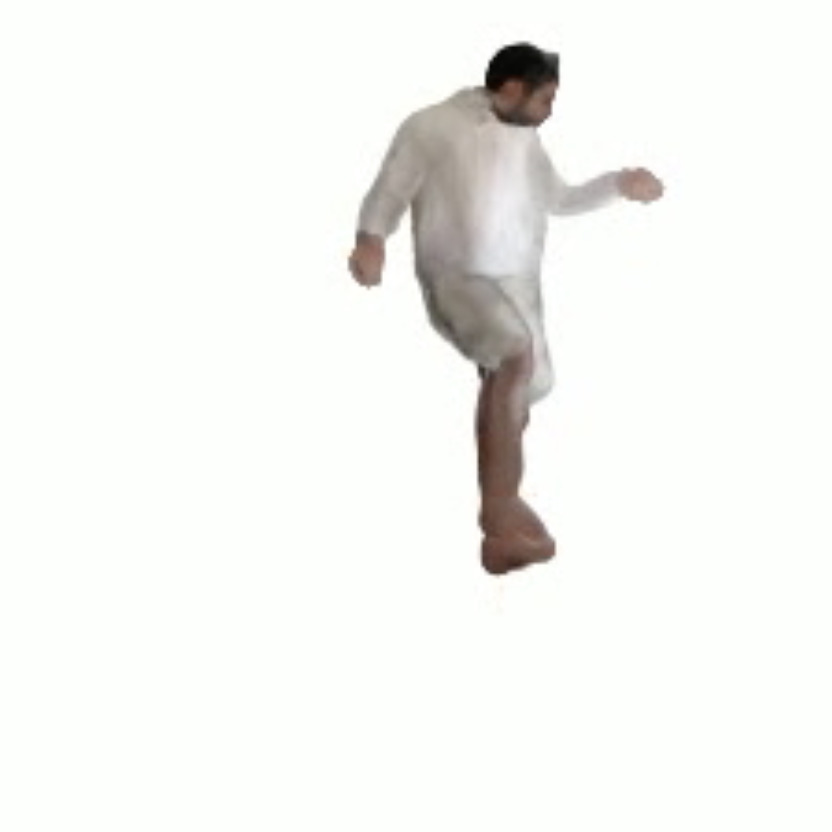}} \\
        \includegraphics[width=0.15\textwidth]{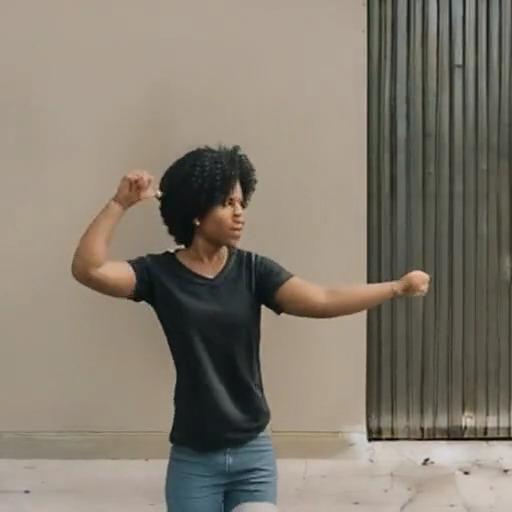} &
        \includegraphics[width=0.15\textwidth]{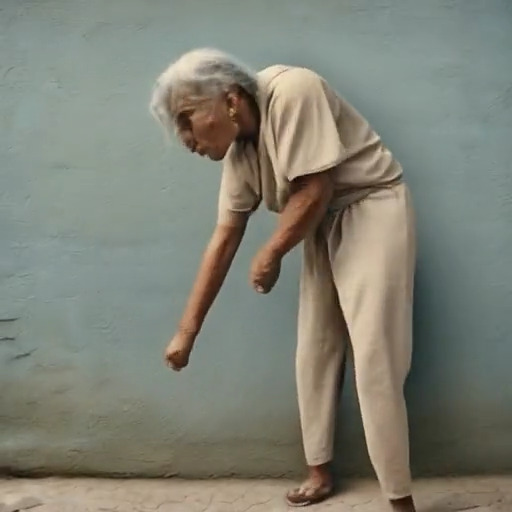} &
        \fbox{\includegraphics[width=0.15\textwidth]{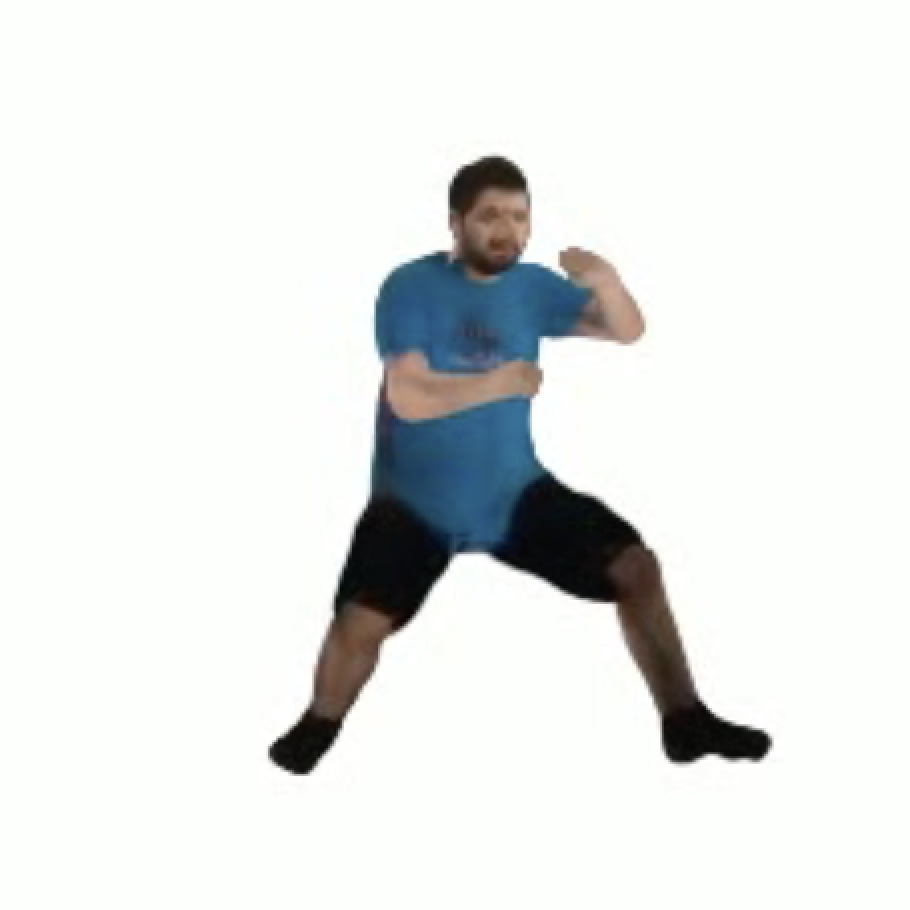}} &
        \includegraphics[width=0.15\textwidth]{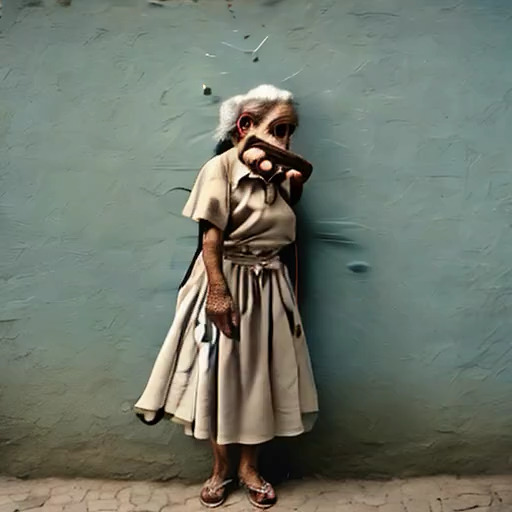} &
        \includegraphics[width=0.15\textwidth]{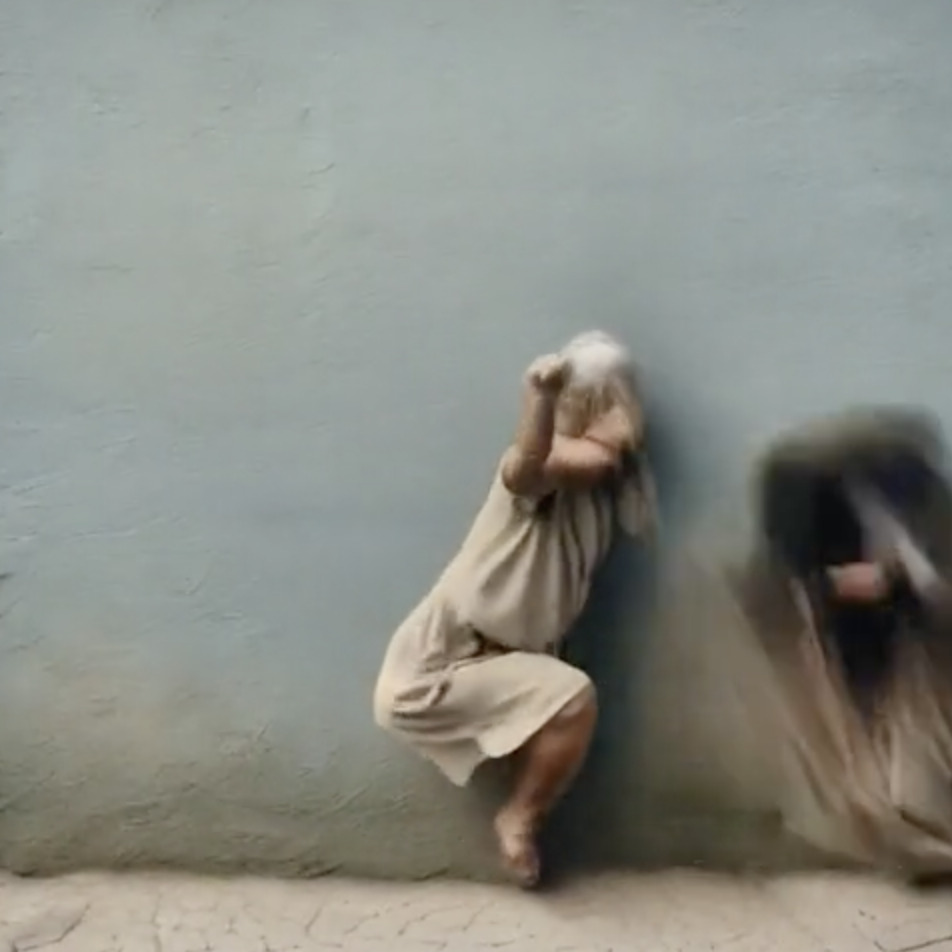} &
        \fbox{\includegraphics[width=0.15\textwidth]{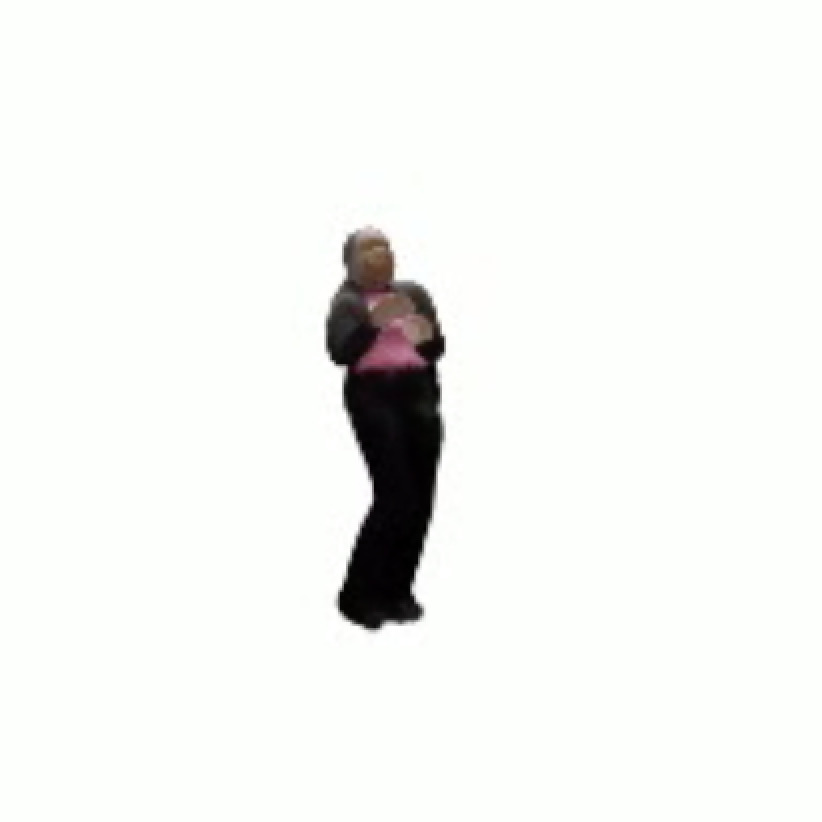}} \\
        \textit{\small AnimateAnyone} & \textit{\small MagicAnimate} & \textit{\small ExAvatar} & \textit{\small AnimateAnyone} & \textit{\small MagicAnimate} & \textit{\small ExAvatar}
    \end{tabular}

    \captionof{figure}{Representative examples of video frames generated by \textit{AnimateAnyone}, \textit{MagicAnimate}, and \textit{ExAvatar}. Shown on the left are frames where the generated human motion is consistent with the reference action video (source) and appears photorealistic; on the right are frames where the generated human motion is not consistent with the reference action video (source) or does not appear photorealistic. \vspace{1em}}
    \label{fig:teaser}
}]

%%%%%%%%% TITLE

%%%%%%%%% ABSTRACT
\begin{abstract}
   Recent pose-transfer methods aim to generate temporally consistent and fully controllable videos of human action where the motion from a reference video is reenacted by a new identity. We evaluate three state-of-the-art pose-transfer methods---\textit{AnimateAnyone}, \textit{MagicAnimate}, and \textit{ExAvatar}---by generating videos with actions and identities outside the training distribution and conducting a participant study about the quality of these videos. In a controlled environment of $20$ distinct human actions, we find that participants, presented with the pose-transferred videos, correctly identify the desired action only $42.92 \%$ of the time. Moreover, the participants find the actions in the generated videos consistent with the reference (source) videos only $36.46 \%$ of the time. These results vary by method: participants find the splatting-based \textit{ExAvatar} more consistent and photorealistic than the diffusion-based \textit{AnimateAnyone} and \textit{MagicAnimate}.
\end{abstract}

%%%%%%%%% BODY TEXT
\maketitle

\section{Introduction}

Pose transfer methods aim to generate images or videos where a novel human identity (target) is shown performing an action consistent with a reference action image or video (source)~\cite{haleem2022human}. Beyond its academic contributions to human body modeling, many applications of this technology have already been identified~\cite{kumar2024pose}: in the entertainment industry, these methods have the potential to enhance human body animation, accelerate production of stunt doubles, and enhance personalization~\cite{zhu2019progressive,jiang2024cinematic}; in healthcare, they are promised to enhance posture therapy~\cite{ekambaram2024real}; and in the fashion industry, these methods will enable virtual try-on where customers shopping online see how a piece of clothing would look on them~\cite{ghodhbani2022you,song2023image}.

Spurred by the potential of the technology, there has been a surge of methods addressing this problem in the literature. In the video pose transfer domain, in particular, the methods fall under two categories based on the underlying architecture: (1) methods based on diffusion (most prominently, \textit{AnimateAnyone}~\cite{hu2024animateanyoneconsistentcontrollable} and \textit{MagicAnimate}~\cite{xu2023magicanimatetemporallyconsistenthuman}) and (2) methods based on 3D Gaussian splatting (most prominently, \textit{ExAvatar}~\cite{moon2024expressive}). 

Promising temporal consistency of the generated identity and high-fidelity control of the movement, these methods achieve state-of-the-art performance across pose transfer benchmarks, including UBC fashion video dataset~\cite{zablotskaia2019dwnet}, TikTok dataset~\cite{jafarian2021learning}, and Ted Talk dataset~\cite{siarohin2021motion}. Yet, despite this promising benchmark performance, the real-world adoption of this technology remains nascent. What's more, human evaluation of these methods remains limited, and the ability of these methods to generalize to identities outside the training distribution is poorly understood.

We, therefore, set out to evaluate the performance of these three state-of-the-art pose transfer methods---\textit{AnimateAnyone}~\cite{hu2024animateanyoneconsistentcontrollable}, \textit{MagicAnimate}~\cite{xu2023magicanimatetemporallyconsistenthuman}, and \textit{ExAvatar}~\cite{moon2024expressive}---outside of benchmarks. To do so, we conducted a participant study focusing on the quality and consistency of videos generated by these methods.

We open this paper by reviewing the development of pose transfer methods up until the methods examined in this paper. We proceed by describing the data and methods used in our generation pipeline. We then shift to the participant study, outlining the methodology and results, and close by discussing the implications of our findings for future research.

\section{Related Work}
\label{sec:related-work}

This section reviews the current state of pose transfer modeling and the evaluation of its out-of-distribution (OOD) performance. Note that, deep learning methods for pose estimation and human understanding have largely paved the way for current pose transfer methods~\cite{kumar2022human}, and so a comprehensive review of these foundational tasks would be granted. Similarly, a discussion of pose transfer applications in downstream tasks could reveal innovative techniques for enhancing performance in specific contexts~\cite{khan2024human,chen2022transfer}. However, such discussions are beyond the scope of this paper, and we thus refer readers to relevant comprehensive surveys~\cite{haleem2022human,zheng2023deep,zhu2023human,dubey2023comprehensive}.

\subsection{Image-level Pose Transfer}

Early deep learning approaches to image-level pose transfer relied on generative adversarial networks (GANs) conditioned on the source image and target pose~\cite{ma2017pose}. Deformable GANs, introduced later, leveraged deformable skip connections to handle large pose variations~\cite{siarohin2018deformable}. The emergence of Transformer-based architecture popularized the use of attention mechanisms in both natural language processing and computer vision domains. Pose transfer methods followed this trend, and so methods that used attention mechanisms to process sequences of poses arrived soon thereafter~\cite{zhu2019progressive}, enabling more detailed and coherent pose transformations. While shared representations of source poses and target human identities had been believed to be crucial for pose transfer, \textit{Wu et al.} showed that disentangling pose and appearance features improves pose transfer results~\cite{wu2023human}, reducing pose ambiguity and appearance inconsistency. Finally, recent work further explored this disentanglement by, for example, permuting image patches to isolate pose from texture~\cite{li2023collecting}.

\subsection{Video-level Pose Transfer}

Pose transfer methods for video generation evolved by adapting image-level architectures to include a temporal dimension. Early approaches, such as Liquid Warping GAN~\cite{liu2019liquid}, introduced a dense 3D flow field that wraps source images, enabling pose transfer across video frames, even from different camera views. The temporal dimension introduced new challenges that had not been addressed by the image-level pose transfer literature, such as pose robustness and cloth dynamics. For instance, \textit{Ren et al.} addressed pose robustness issues through pose augmentations in a two-stage network~\cite{ren2020human}, while \textit{Kappel et al.} tackled cloth dynamics by incorporating clothing-specific modules into the architecture and new training objectives~\cite{kappel2021high}. Hybrid approaches combining deep learning with computer graphics concepts have also been explored~\cite{yu2023bidirectionally}.

The recent boom in generative AI has led to diffusion-based methods applied for video-level pose transfer. Early diffusion-based approaches to video-level pose transfer relied on fixed seeds or CLIP embeddings~\cite{radford2021learningtransferablevisualmodels, karras2023dreamposefashionimagetovideosynthesis}. These methods struggled with temporal consistency and artifacts like flickering~\cite{wang2023edittemporalconsistentvideosimage, roy2022multiscaleattentionguidedpose, ren2020humanmotiontransferposes, LIU2021107024}. \textit{MagicAnimate}~\cite{xu2023magicanimatetemporallyconsistenthuman} addressed some of these issues by employing temporal attention blocks, an appearance encoder, and a video fusion technique. However, its reliance on DensePose representations~\cite{güler2018denseposedensehumanpose} limited the transfer quality of fine details such as fingers. \textit{AnimateAnyone}~\cite{hu2024animateanyoneconsistentcontrollable} improved upon this by using skeletal representations instead of DensePose, enabling the transfer of finer detail and more stability. Still, some issues with temporal consistency and poor generalization to novel environments persisted. The latest methods have thus aimed to combine generative AI techniques with concepts from computer graphics and address these challenges. Most prominently, \textit{ExAvatar}~\cite{moon2024expressive} leverages an expressive whole-body 3D Gaussian representation built on top of an ensemble of body representations. By benchmark performance, \textit{MagicAnimate}, \textit{AnimateAnyone}, and \textit{ExAvatar} represent the state of the art in video-level pose transfer.

\subsection{Out-of-Domain Evaluation}

The scholarship examining performance of pose transfer methods under OOD conditions remains limited. One study explored the OOD performance of image-level pose transfer by analyzing the datasets used for model training and inference~\cite{chen2023open}. Other research has examined OOD performance of pose estimation methods~\cite{liu2023poseexaminer,shukla2022vl4pose}. However, a significant gap exists in the literature regarding the OOD evaluation of video-level pose transfer. Moreover, no recent studies have gone beyond benchmark scores to assess these systems through human participant evaluations.

\begin{figure}[t]
    \centering
    \begin{tabular}{c@{\hskip 4pt}c@{\hskip 4pt}c}
        \includegraphics[width=0.31\linewidth]{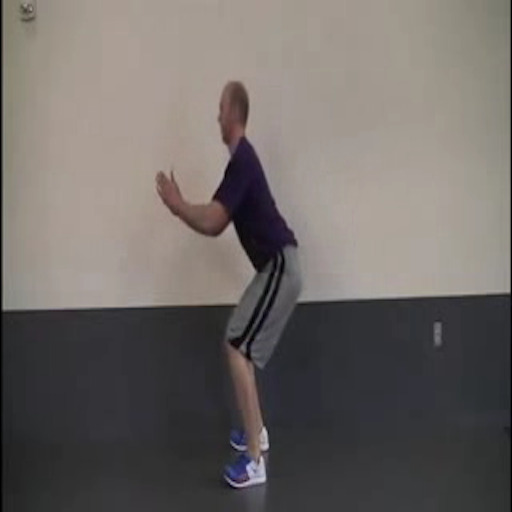} &
        \includegraphics[width=0.31\linewidth]{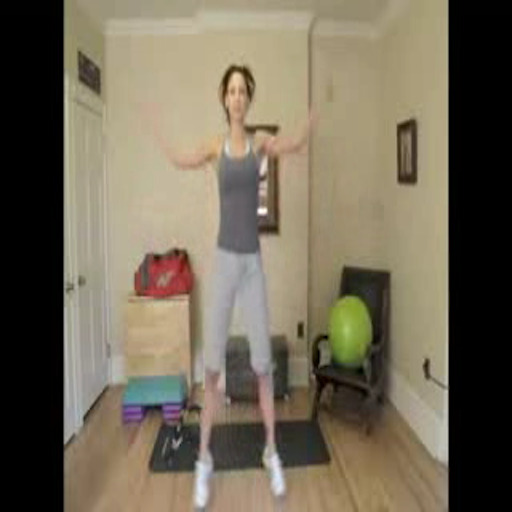} &
        \includegraphics[width=0.31\linewidth]{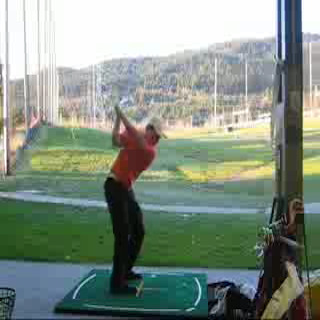} \\
        \small (a) Bodyweight squats & \small (b) Jumping jack & \small (c) Golf swing
    \end{tabular}
    \caption{Representative examples of frames from UCF101, cropped to fit the layout.}
    \label{fig:ucf101-videos}
\end{figure}

\begin{figure}[t]
    \centering
    \begin{tabular}{c@{\hskip 4pt}c@{\hskip 4pt}c}
        \includegraphics[width=0.31\linewidth]{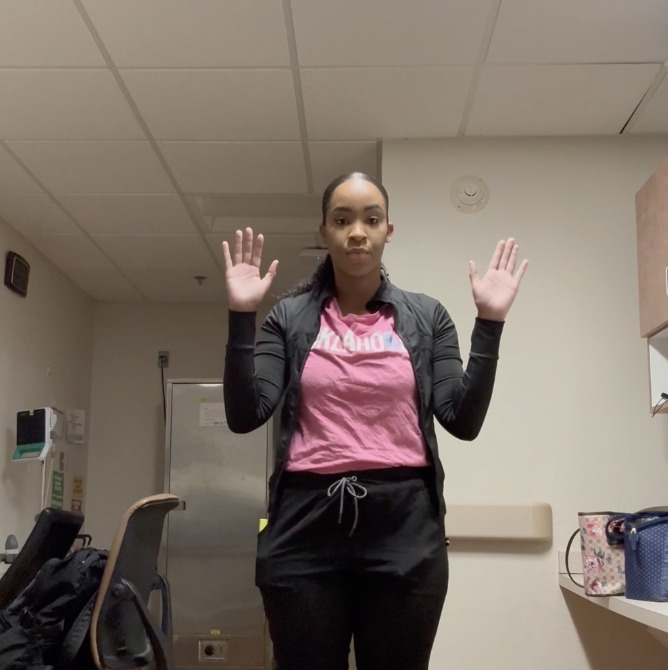} &
        \includegraphics[width=0.31\linewidth]{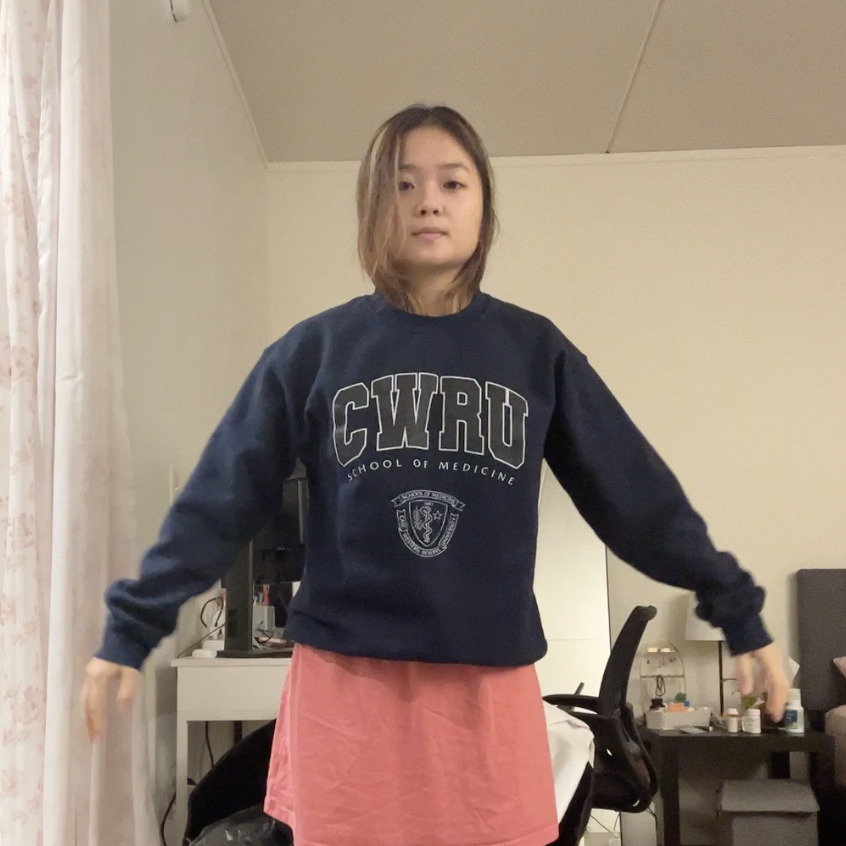} &
        \includegraphics[width=0.31\linewidth]{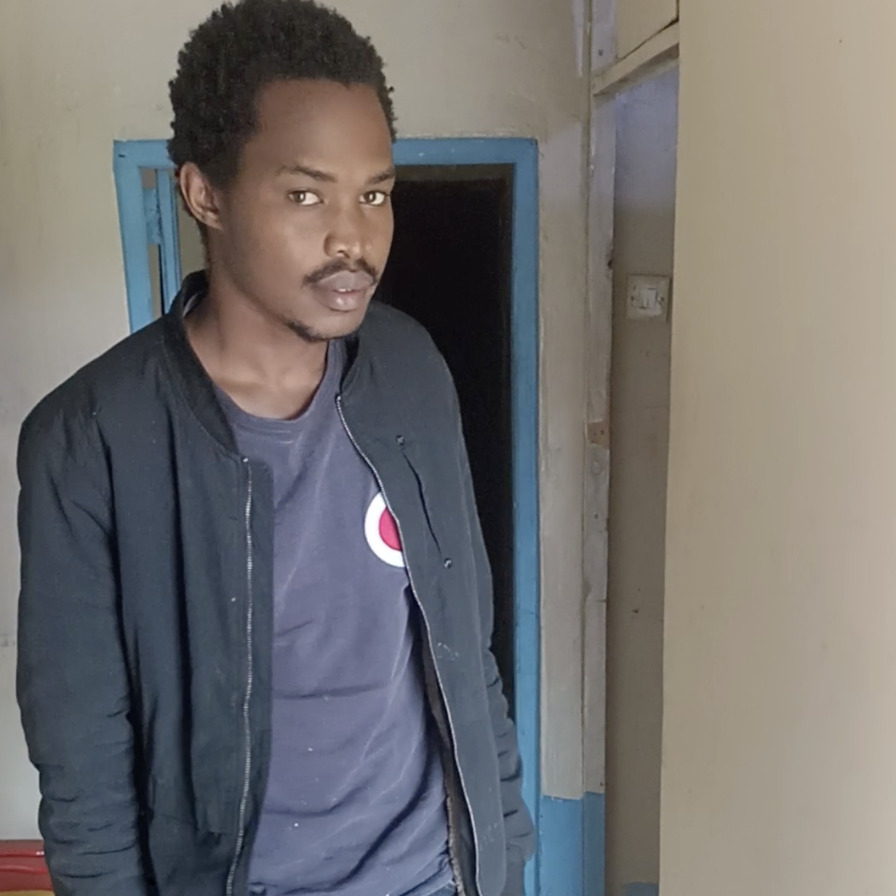} 
    \end{tabular}
    \caption{Representative examples of video frames from the RANDOM People dataset, cropped to fit the layout.}
    \label{fig:random-people-videos}
\end{figure}

\begin{figure}[t]
    \centering
    \begin{tabular}{c@{\hskip 4pt}c@{\hskip 4pt}c}
        \includegraphics[width=0.31\linewidth]{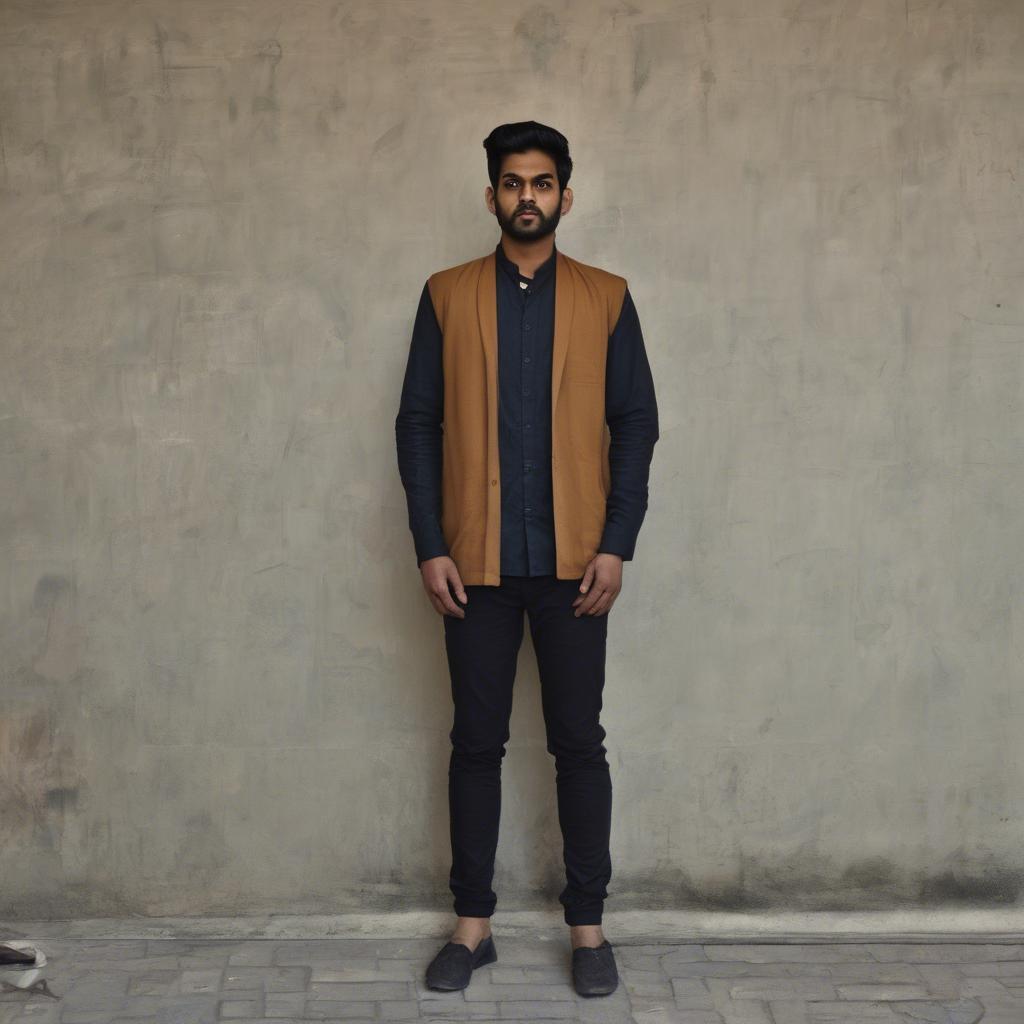} &
        \includegraphics[width=0.31\linewidth]{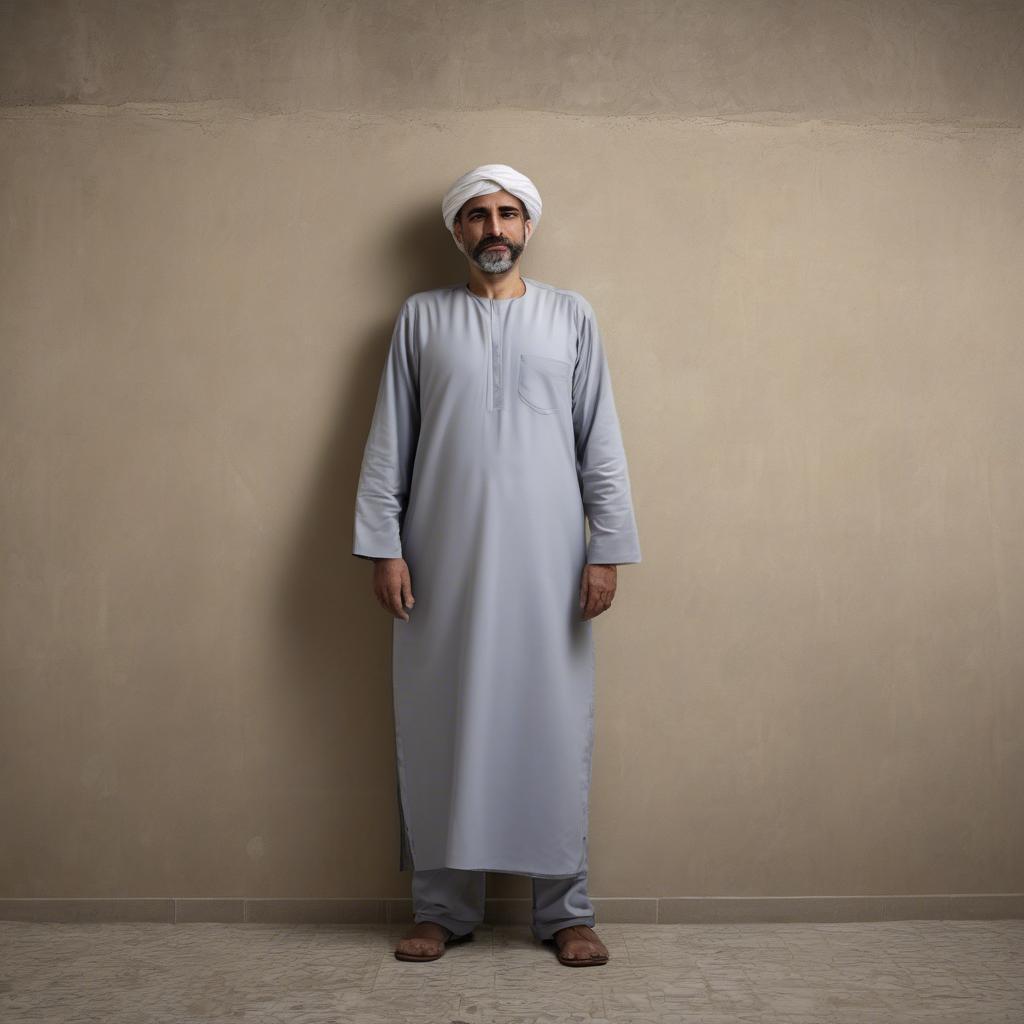} &
        \includegraphics[width=0.31\linewidth]{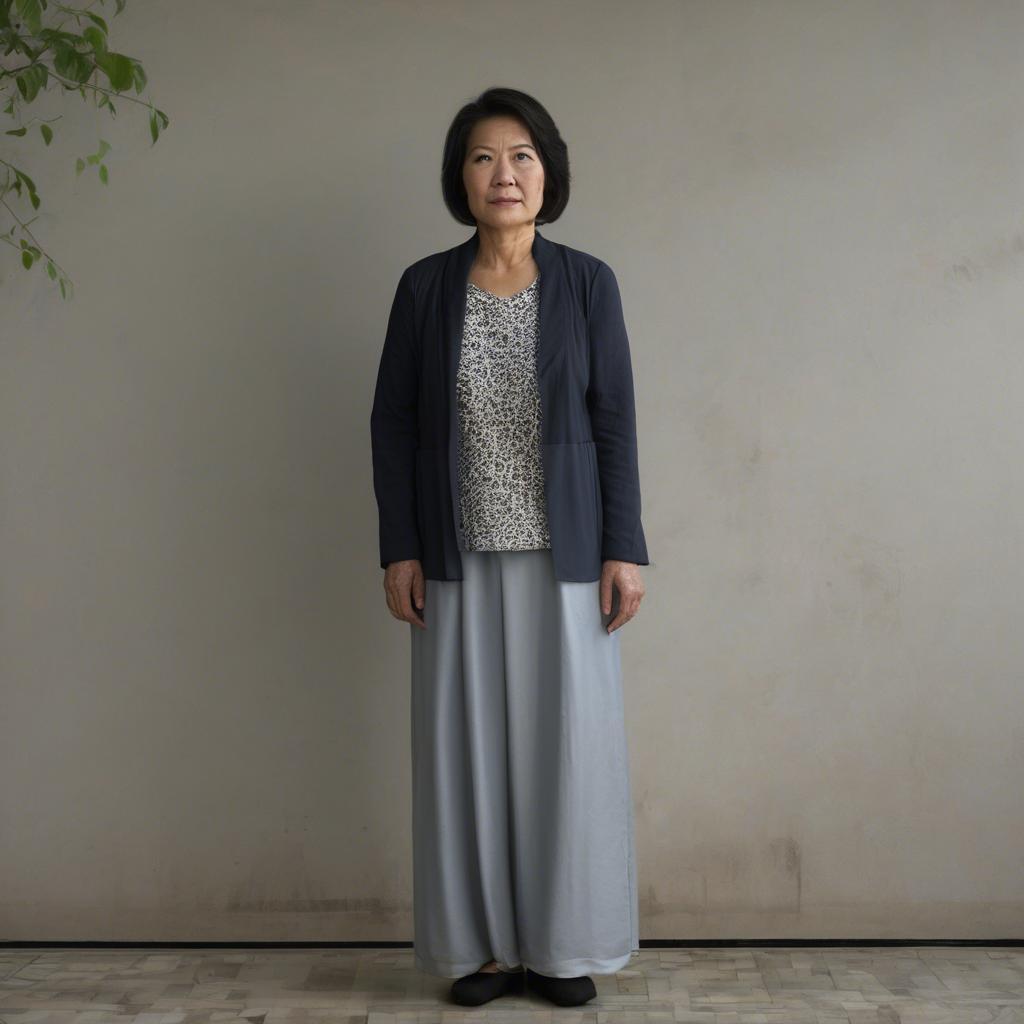} \\
    \end{tabular}
    \caption{Representative examples of target identities.}
    \label{fig:identities}
\end{figure}

\section{Data and Methods}
\label{sec:methods}

This section outlines how we generated pose-transferred videos for our study. For each pose transfer method examined---\textit{AnimateAnyone}~\cite{hu2024animateanyoneconsistentcontrollable}, \textit{MagicAnimate}~\cite{xu2023magicanimatetemporallyconsistenthuman}, and \textit{ExAvatar}~\cite{moon2024expressive}---the data preparation involved three steps: (1) collecting reference action videos (see Section~\ref{subsec:reference-action-videos}), (2) collecting novel human identities (see Section~\ref{subsec:novel-human-identities}), and (3) generating new videos using pose transfer (see Section~\ref{subsec:pose-transfer}). Step 1 was consistent across all methods; Steps 2 and 3 varied depending on the specific pose transfer method.

For each pose transfer method, a total of $840$ videos were generated ($42$ videos per class). From these, $22$ videos were later selected for the survey: $10$ videos for Task 1, another $10$ videos for Task 2, and $2$ videos for Task 3 (see Section~\ref{sec:survey-methodology} for survey details). The reference action videos and novel human identities were sampled at random while avoiding repetition and ensuring that all activities are present at least once for each pose transfer method. Representative examples of frames from these videos are shown in Figure~\ref{fig:teaser}. To enhance reproducibility of our work, we open-source these videos at \url{https://github.com/matyasbohacek/pose-transfer-human-motion}.

\subsection{Reference Action Videos}
\label{subsec:reference-action-videos}

A subset of $60$ videos spanning $20$ classes from the UCF101~\cite{soomro2012ucf101dataset101human} human action video dataset was used as our reference action (source) video set. UCF101~\cite{soomro2012ucf101dataset101human} is a large-scale action recognition dataset comprising $101$ action classes, more than $13,000$ videos, and $27$ hours of footage. The dataset, scraped from YouTube in 2012, consists of predominantly low-resolution videos. Representative examples of frames from the dataset are shown in Figure~\ref{fig:ucf101-videos}.
For this study, we selected $20$ classes representing actions performed by a single person with minimal or no use of external objects (see Appendix~\ref{app:selected-action-classes} for the complete list of selected action classes). We manually selected three videos per class.

\begin{figure}[t]
    \centering
    \includegraphics[width=0.97\linewidth]{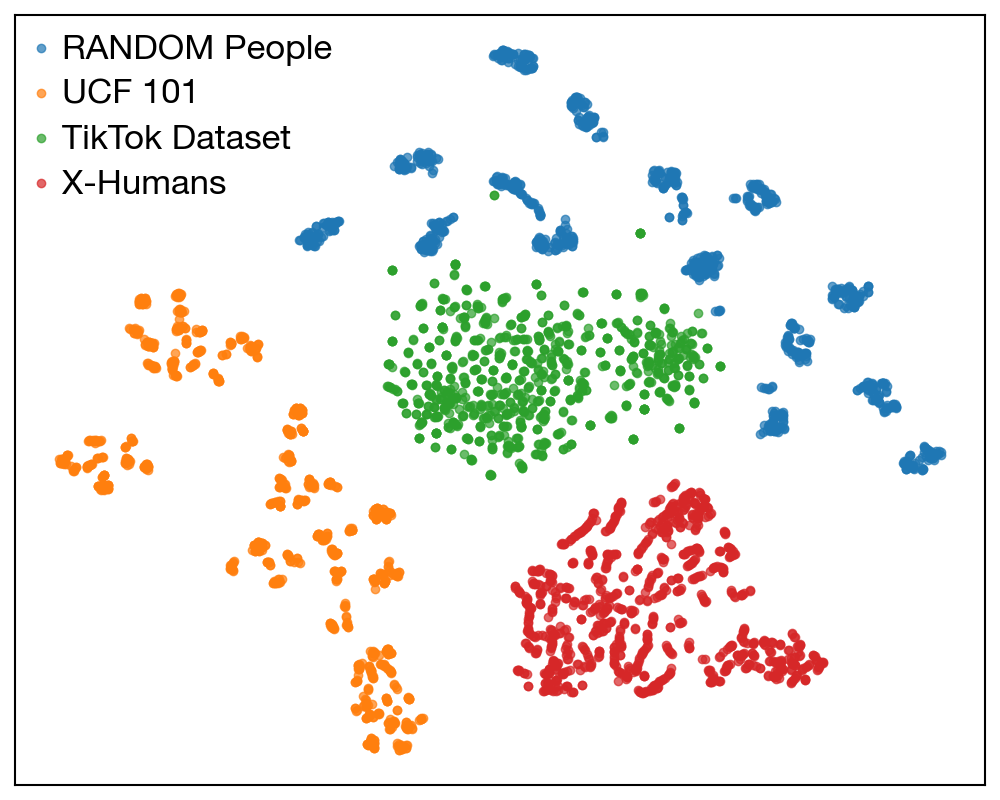}
    \vspace{0.35cm}
    \caption{t-SNE visualization of CLIP embeddings representing datasets used in the OOD evaluation (UCF101 and RANDOM People) and datasets used to train the evaluated pose transfer methods (TikTok Dataset and X-Humans).}
    \label{fig:tsne}
\end{figure}

\begin{figure*}[t]
    \centering
    \begin{tabular}{c@{\hskip 4pt}c@{\hskip 4pt}c@{\hskip 25pt}c@{\hskip 4pt}c@{\hskip 4pt}c}
        \includegraphics[width=0.15\textwidth]{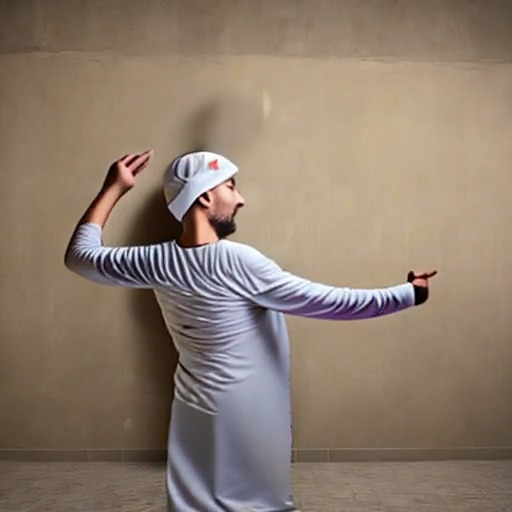} &
        \includegraphics[width=0.15\textwidth]{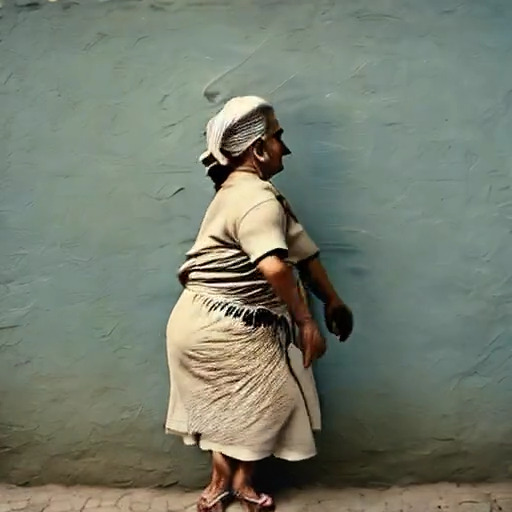} &
        \includegraphics[width=0.15\textwidth]{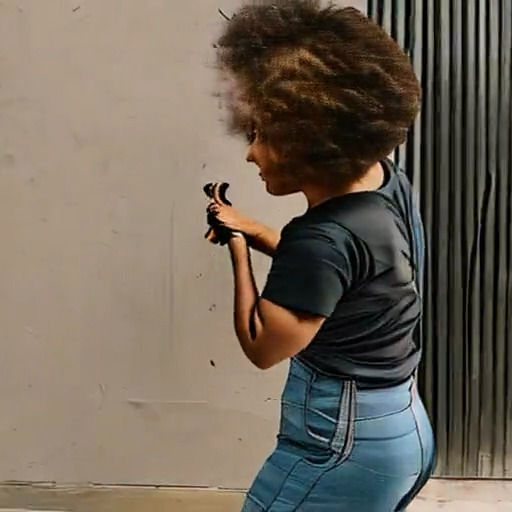} &
        \includegraphics[width=0.15\textwidth]{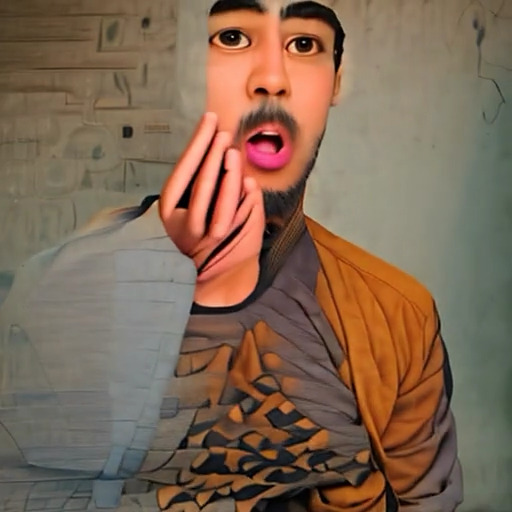} &
        \includegraphics[width=0.15\textwidth]{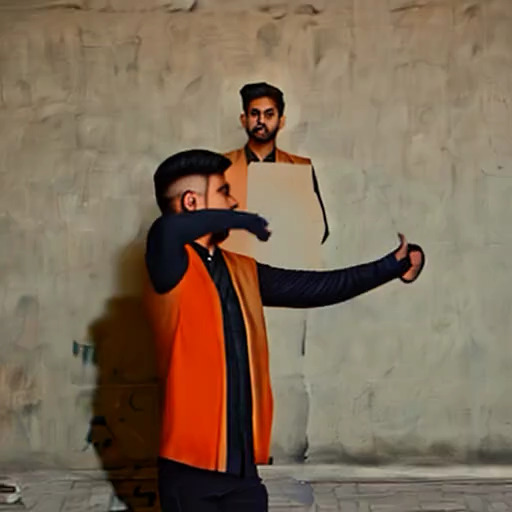} &
        \includegraphics[width=0.15\textwidth]{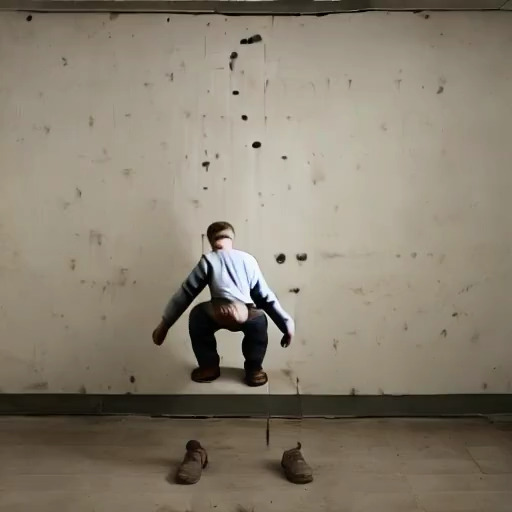} \\

        \includegraphics[width=0.15\textwidth]{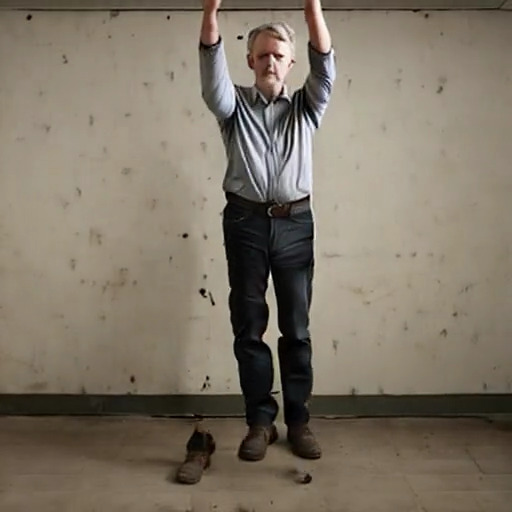} &
        \includegraphics[width=0.15\textwidth]{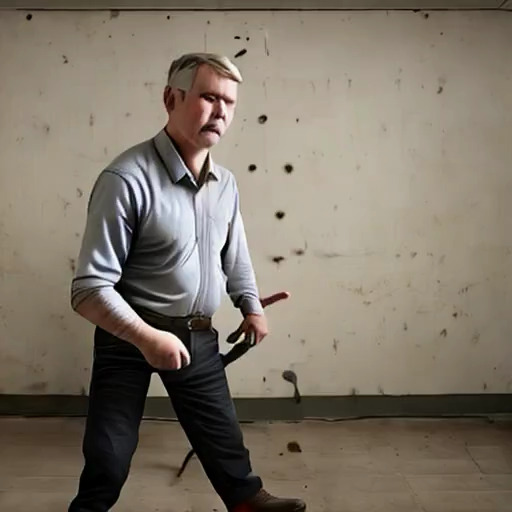} &
        \includegraphics[width=0.15\textwidth]{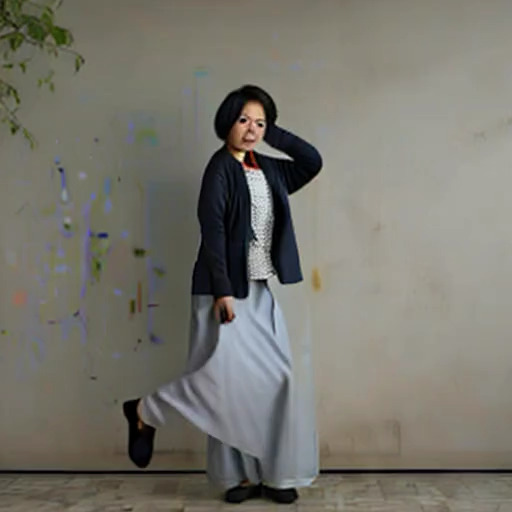} &
        \includegraphics[width=0.15\textwidth]{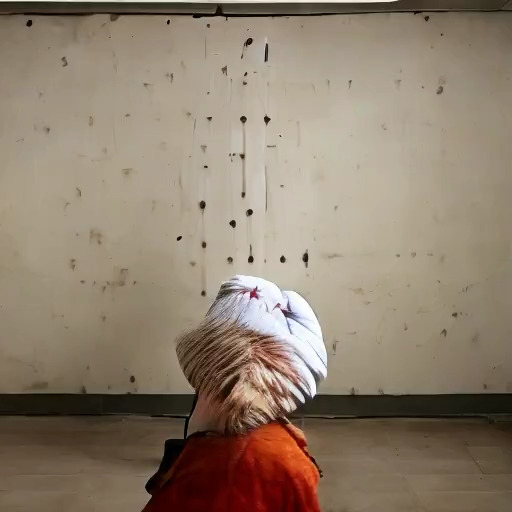} &
        \includegraphics[width=0.15\textwidth]{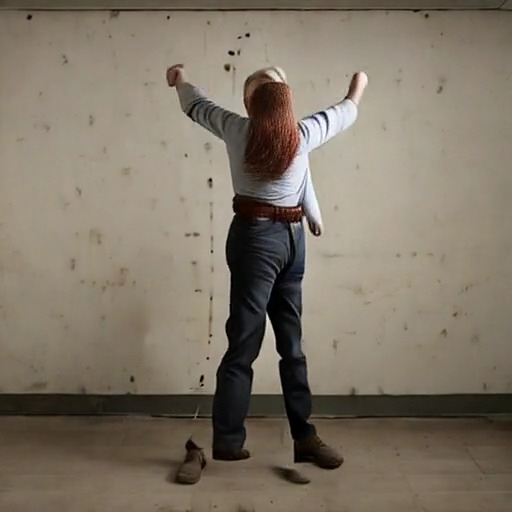} &
        \includegraphics[width=0.15\textwidth]{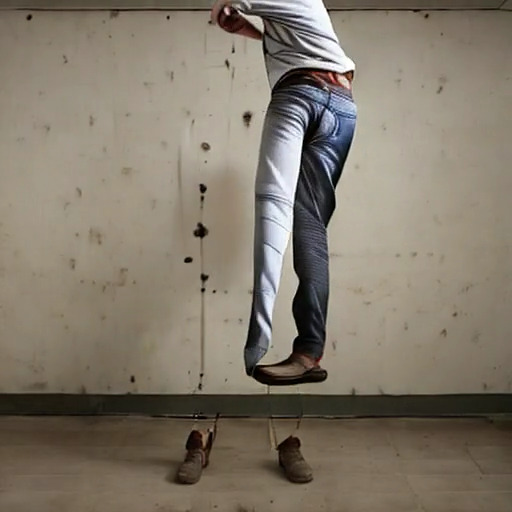} \\

        \includegraphics[width=0.15\textwidth]{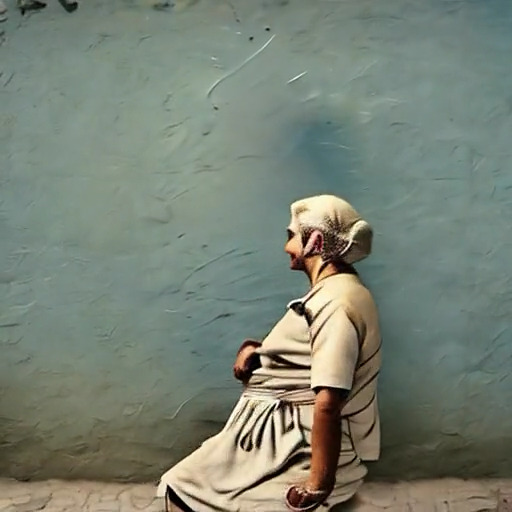} &
        \includegraphics[width=0.15\textwidth]{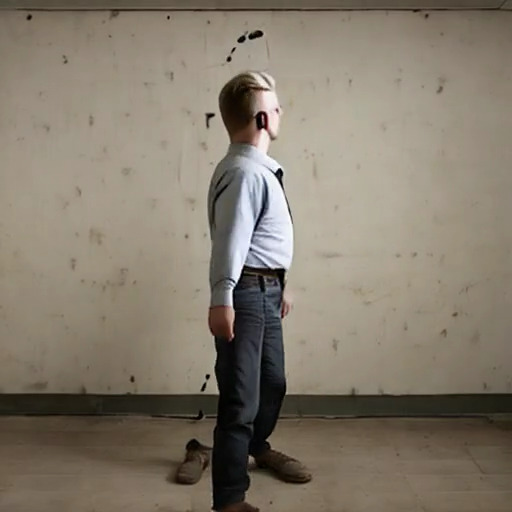} &
        \includegraphics[width=0.15\textwidth]{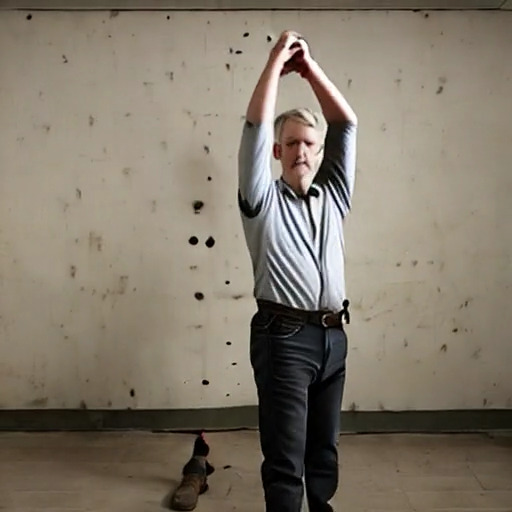} &
        \includegraphics[width=0.15\textwidth]{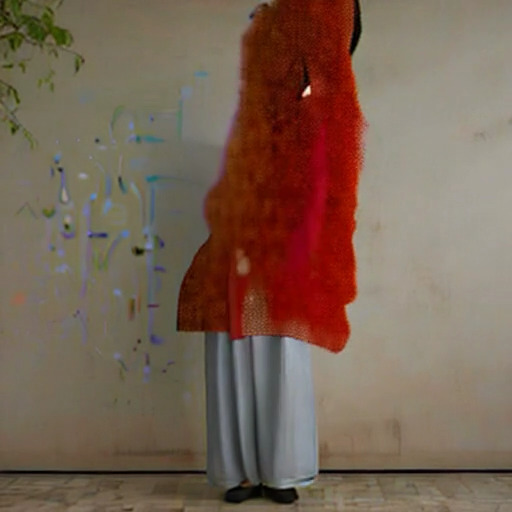} &
        \includegraphics[width=0.15\textwidth]{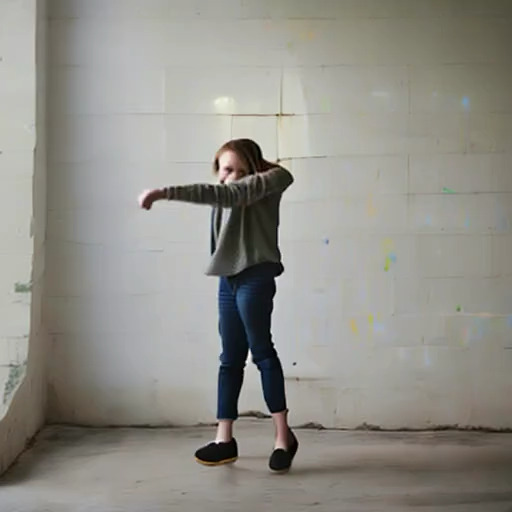} &
        \includegraphics[width=0.15\textwidth]{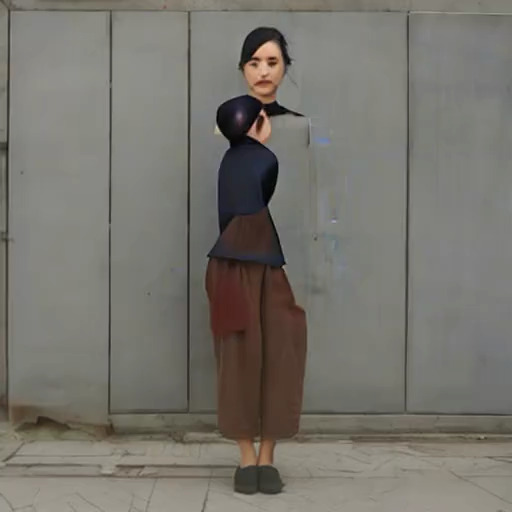} \\

        \multicolumn{3}{c}{\small Consistent and photorealistic frames} \hspace{26pt} &
        \multicolumn{3}{c}{\small Inconsistent or non-photorealistic frames} 

    \end{tabular}
    \caption{Representative examples of frames generated by \textit{AnimateAnyone}.}
    \label{fig:examples-animateanyone}
\end{figure*}

\subsection{Novel Human Identities}
\label{subsec:novel-human-identities}

To specify the novel human identity (target), the diffusion-based methods employed in our study, \textit{AnimateAnyone} and \textit{MagicAnimate}, expect a static image of the identity. The splatting-based method, \textit{ExAvatar}, on the other hand, expects a video of the identity. We, thus, describe the method used for each type of pose transfer separately.

\textbf{AnimateAnyone and MagicAnimate.} AI-generated images of $14$ individuals were used as the novel human identity (target) set for \textit{AnimateAnyone} and \textit{MagicAnimate}. These images were generated using Stable Diffusion XL~\cite{podell2023sdxl} text-to-image model with the prompt "\textit{[age] [ethnicity] [sex] standing in front of a wall, full body and hands visible}". The age descriptions were: "\textit{young}", "\textit{middle-aged}", or "\textit{old}". The ethnicity descriptions were: "\textit{Black}", "\textit{Hispanic}", "\textit{East Asian}", "\textit{South Asian}", or "\textit{white}". The sex descriptions were: "\textit{man}" or "\textit{woman}". Representative examples of these images are shown in Figure~\ref{fig:identities}. We ensured that this set was diverse in age, gender, and ethnicity.

\textbf{ExAvatar.} Videos of $15$ individuals from the RANDOM People dataset\footnote{\url{https://anonymous.4open.science/r/random-people-dataset-D70F/}} diverse in age, gender, and ethnicity were used as the novel human identity (target) set for \textit{ExAvatar}. Representative examples of these images are shown in Figure~\ref{fig:random-people-videos}.

\begin{figure*}[t]
    \centering
    \begin{tabular}{c@{\hskip 4pt}c@{\hskip 4pt}c@{\hskip 25pt}c@{\hskip 4pt}c@{\hskip 4pt}c}
        \includegraphics[width=0.15\textwidth]{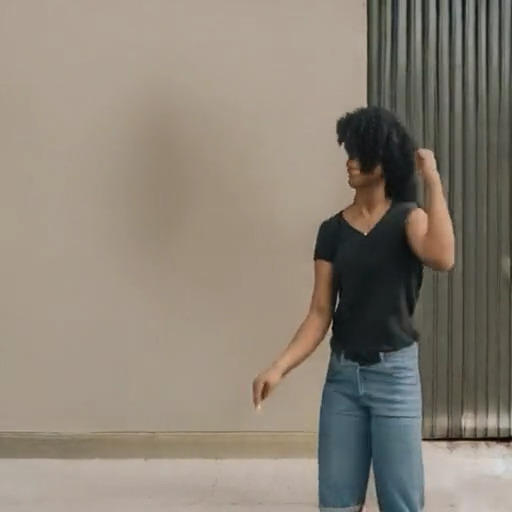} &
        \includegraphics[width=0.15\textwidth]{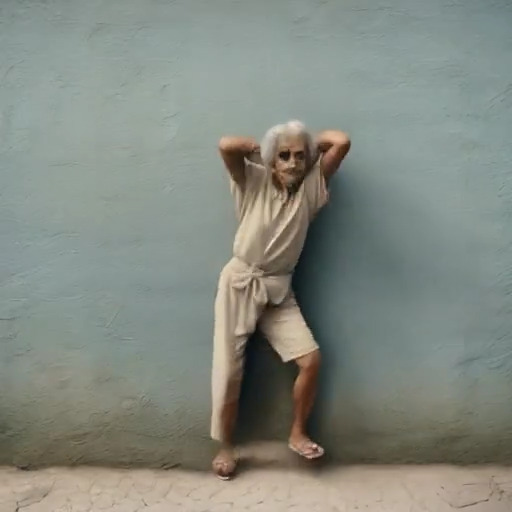} &
        \includegraphics[width=0.15\textwidth]{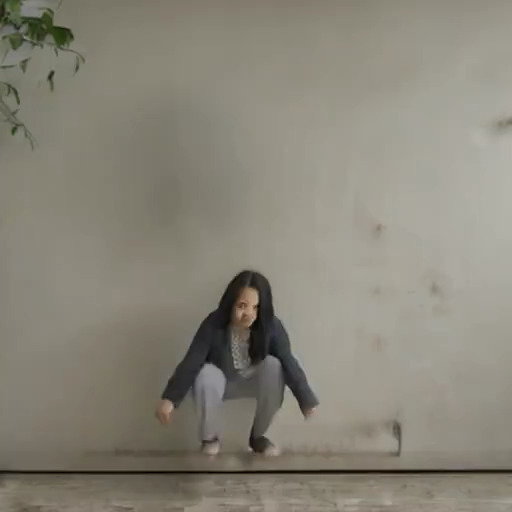} &
        \includegraphics[width=0.15\textwidth]{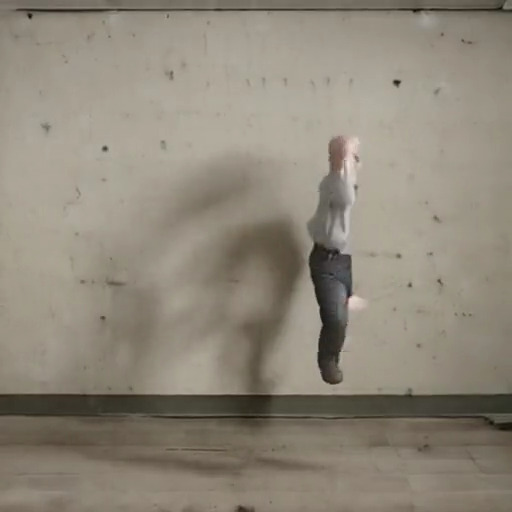} &
        \includegraphics[width=0.15\textwidth]{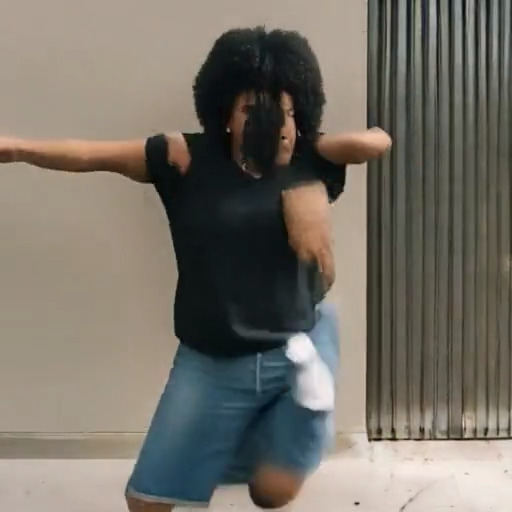} &
        \includegraphics[width=0.15\textwidth]{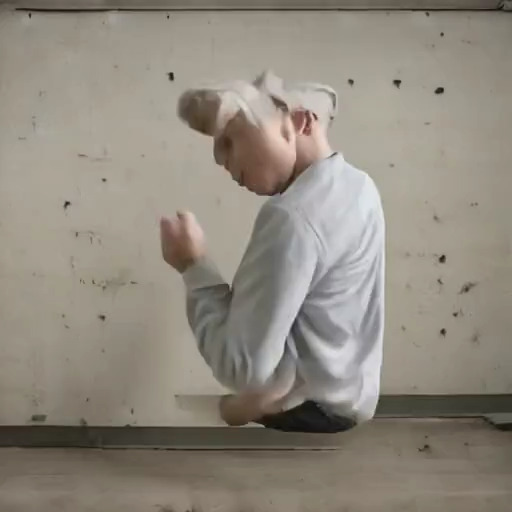} \\

        \includegraphics[width=0.15\textwidth]{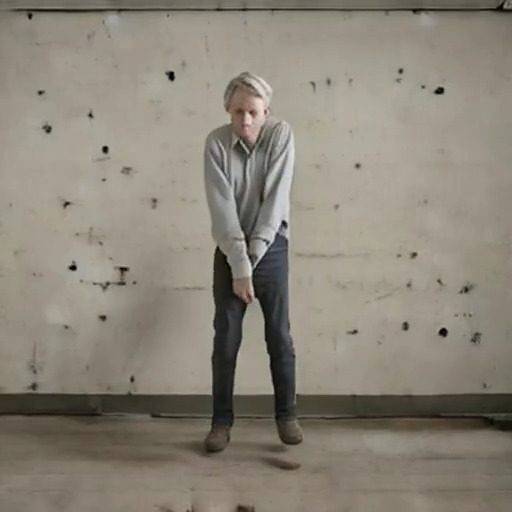} &
        \includegraphics[width=0.15\textwidth]{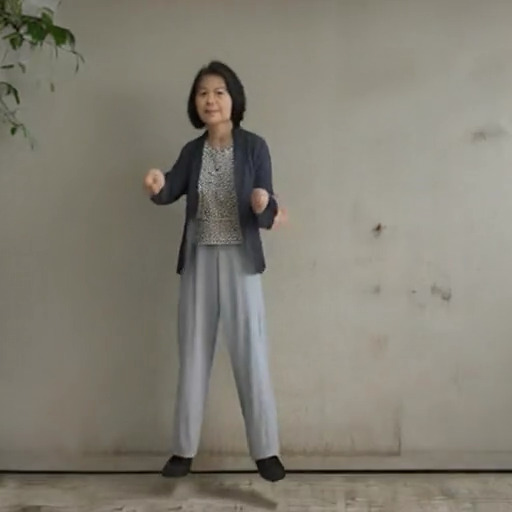} &
        \includegraphics[width=0.15\textwidth]{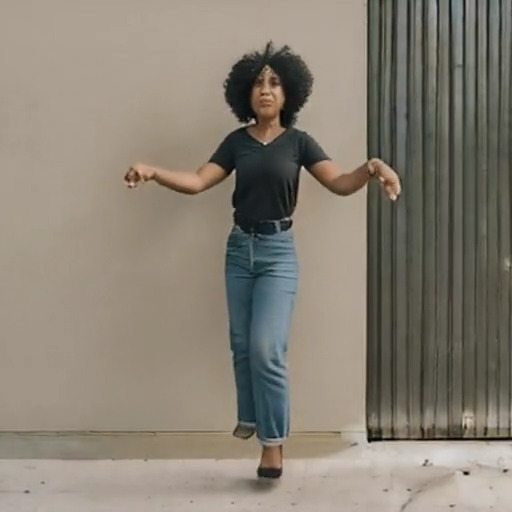} &
        \includegraphics[width=0.15\textwidth]{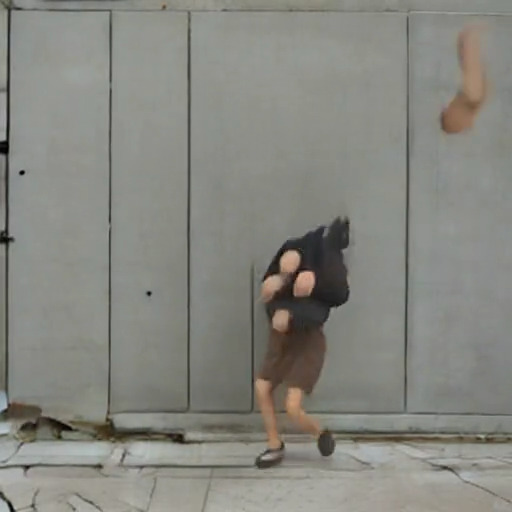} &
        \includegraphics[width=0.15\textwidth]{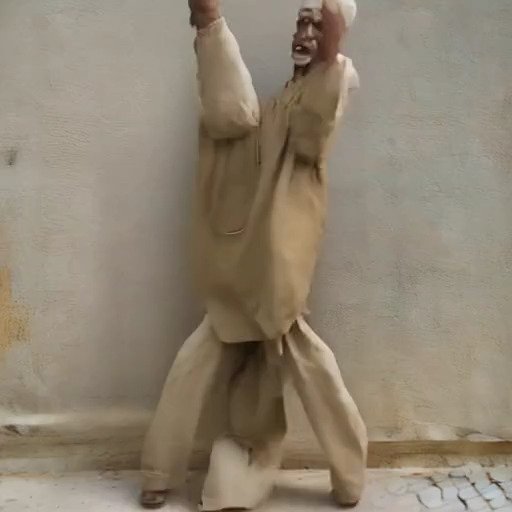} &
        \includegraphics[width=0.15\textwidth]{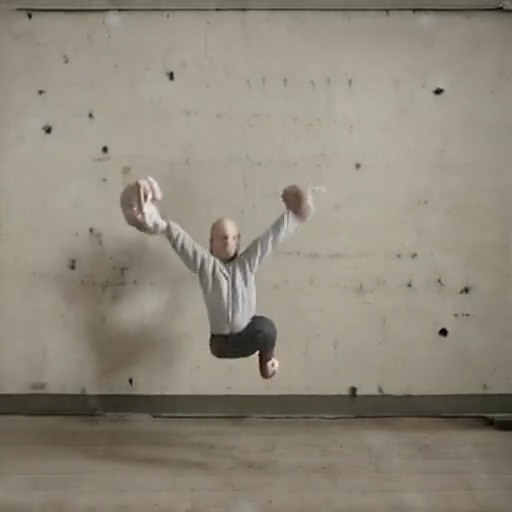} \\
        
        \includegraphics[width=0.15\textwidth]{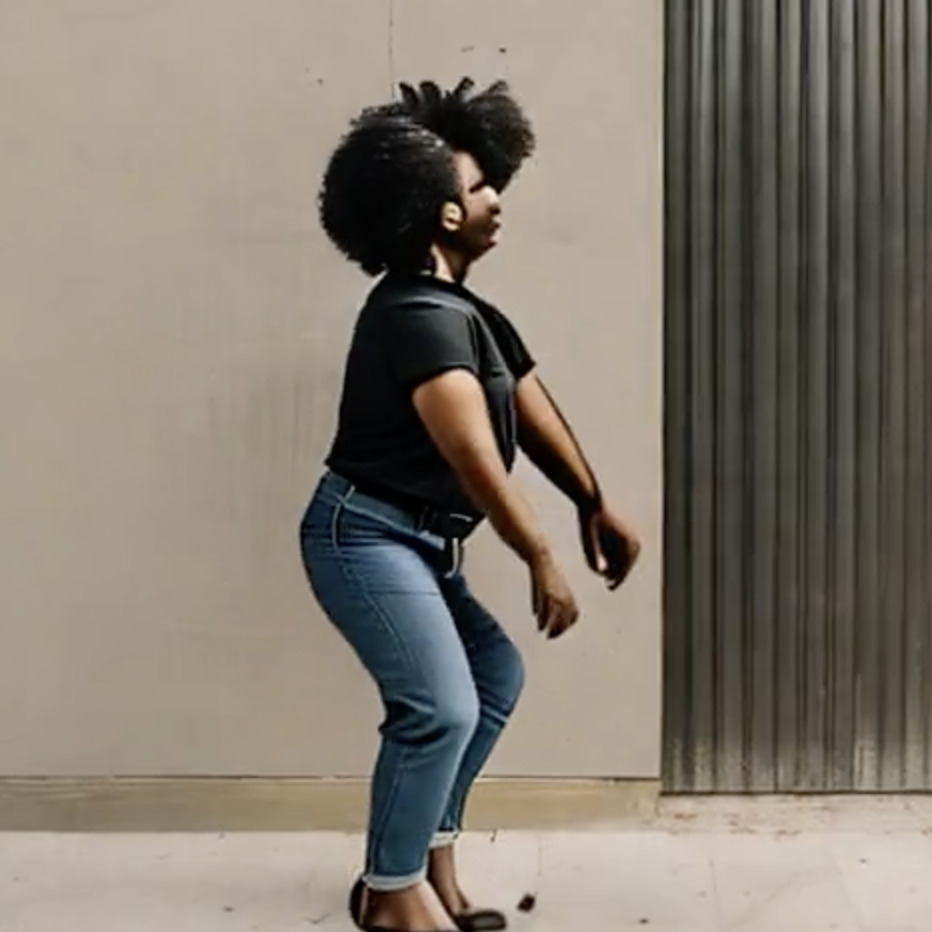} &
        \includegraphics[width=0.15\textwidth]{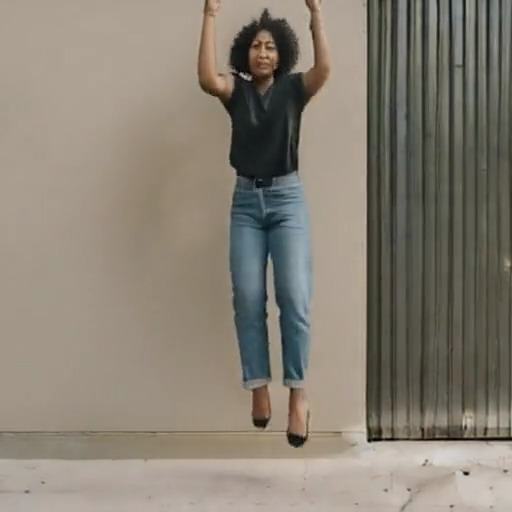} &
        \includegraphics[width=0.15\textwidth]{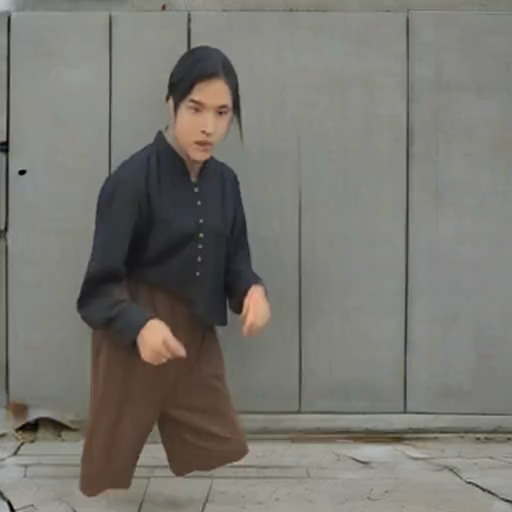} &
        \includegraphics[width=0.15\textwidth]{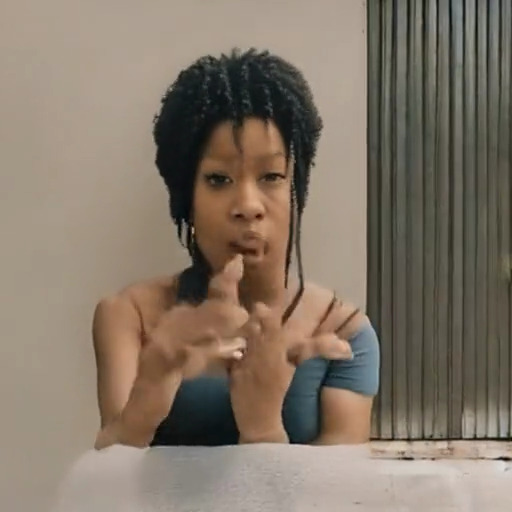} &
        \includegraphics[width=0.15\textwidth]{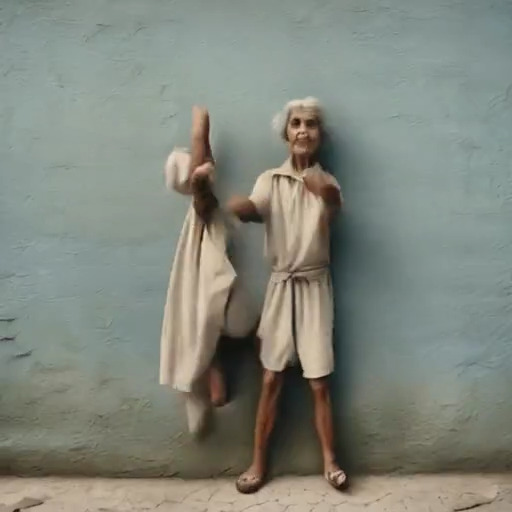} &
        \includegraphics[width=0.15\textwidth]{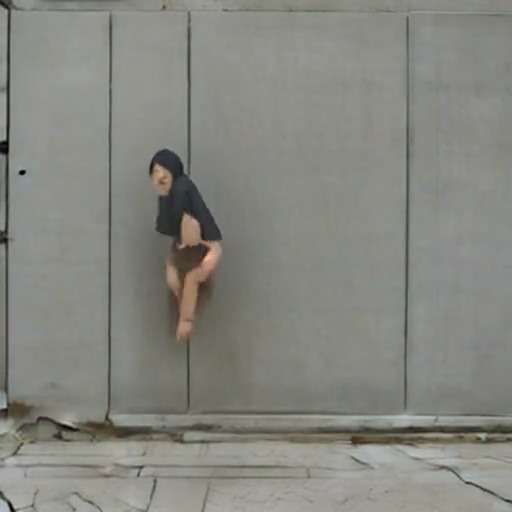} \\

        \multicolumn{3}{c}{\small Consistent and photorealistic frames} \hspace{26pt} &
        \multicolumn{3}{c}{\small Inconsistent or non-photorealistic frames} 
    \end{tabular}
    \caption{Representative examples of video frames generated by \textit{MagicAnimate}.}
    \label{fig:examples-magicanimate}
\end{figure*}

\subsection{Distribution Analysis}

To ensure the suitability of the employed datasets--UCF101 and RANDOM People---for an OOD evaluation, we performed a t-SNE analysis~\cite{hinton2008visualizing} comparing these datasets to the training datasets of the evaluated pose transfer methods. Specifically, we compared them against the TikTok Dataset~\cite{jafarian2021learning}, used to train \textit{MagicAnimate}, and the X-Humans Dataset~\cite{shen2023x}, used to train \textit{ExAvatar}. The dataset used to train \textit{AnimateAnyone} was not disclosed by its authors.

For each dataset, $1,000$ video frames were chosen at random and converted into CLIP embeddings~\cite{radford2021learning} using the ViT-B $32$-patch model variant\footnote{\url{https://huggingface.co/openai/clip-vit-base-patch32}}. The t-SNE visualization of these embeddings, shown in Figure~\ref{fig:tsne}, was created using the scikit-learn library~\cite{pedregosa2011scikit} with default parameters. The visualization demonstrates a clear separation among all datasets.

\subsection{Pose Transfer} 
\label{subsec:pose-transfer}

Given a reference action video (source) and a novel human identity image or video (target), each examined pose transfer method generated a video in which the novel human identity reenacted the human action shown in the reference action video.

\textbf{AnimateAnyone. } This method represents the novel human identity as an image (in our case, sourced by SD XL images) and integrates ReferenceNet within its diffusion-based architecture to enhance temporal consistency of the novel human identity (target). Due to the absence of an official implementation, an open-source implementation of \textit{AnimateAnyone}\footnote{\url{https://github.com/MooreThreads/Moore-AnimateAnyone/}}, which has been shown to have on-par or better performance, was used. Examples of representative video frames generated by \textit{AnimateAnyone} are shown in Figure~\ref{fig:examples-animateanyone}.

\textbf{MagicAnimate. } Also representing the novel human identity as an image (in our case, sourced by SD XL images), \textit{MagicAnimate} employs temporal attention blocks and an appearance encoder within its diffusion-based architecture to improve temporal consistency, fidelity, and smoothness. The official implementation\footnote{\url{https://github.com/magic-research/magic-animate/}} was used. Examples of representative video frames generated by \textit{MagicAnimate} are shown in Figure~\ref{fig:examples-magicanimate}.

\begin{figure*}[t]
    \centering
    \begin{tabular}{c@{\hskip 4pt}c@{\hskip 4pt}c@{\hskip 25pt}c@{\hskip 4pt}c@{\hskip 4pt}c}
        \fbox{\includegraphics[width=0.15\textwidth]{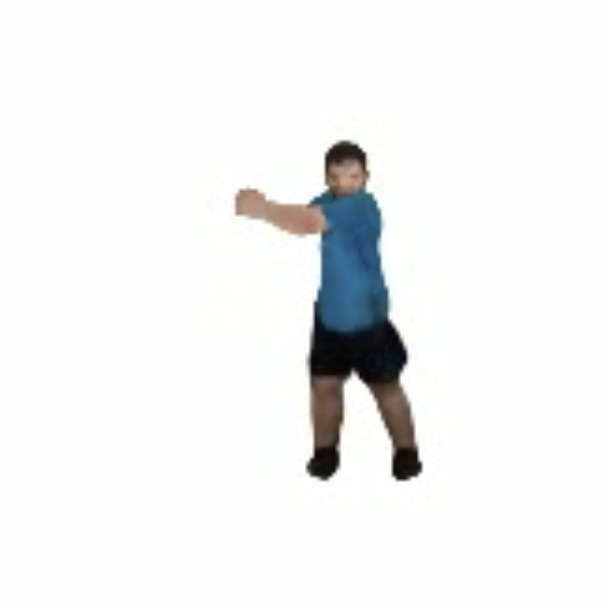}} &
        \fbox{\includegraphics[width=0.15\textwidth]{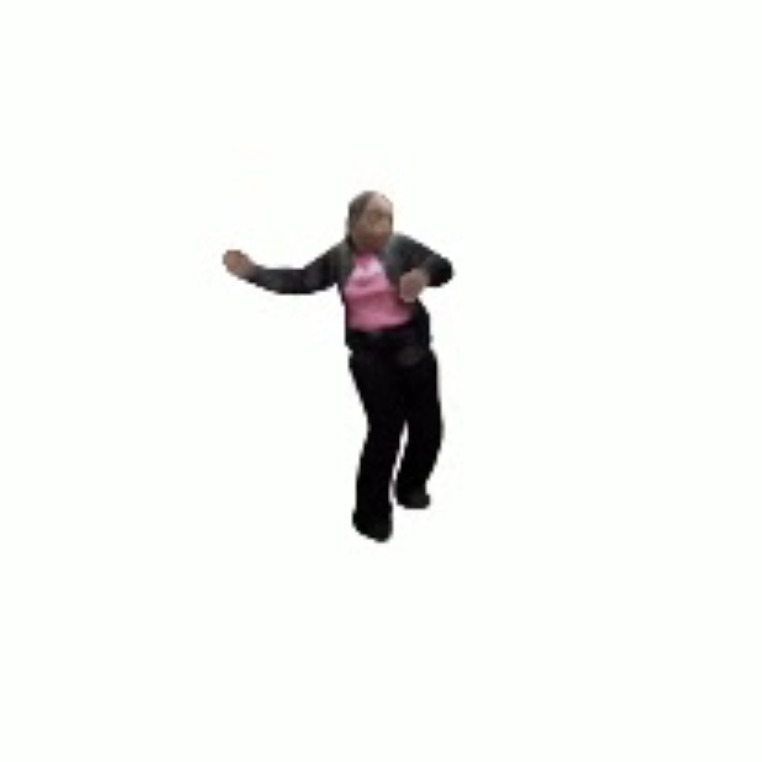}} &
        \fbox{\includegraphics[width=0.15\textwidth]{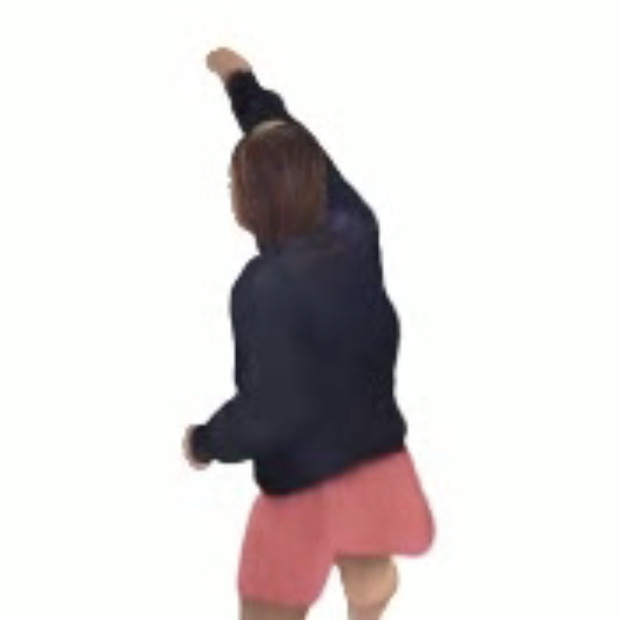}} &
        \fbox{\includegraphics[width=0.15\textwidth]{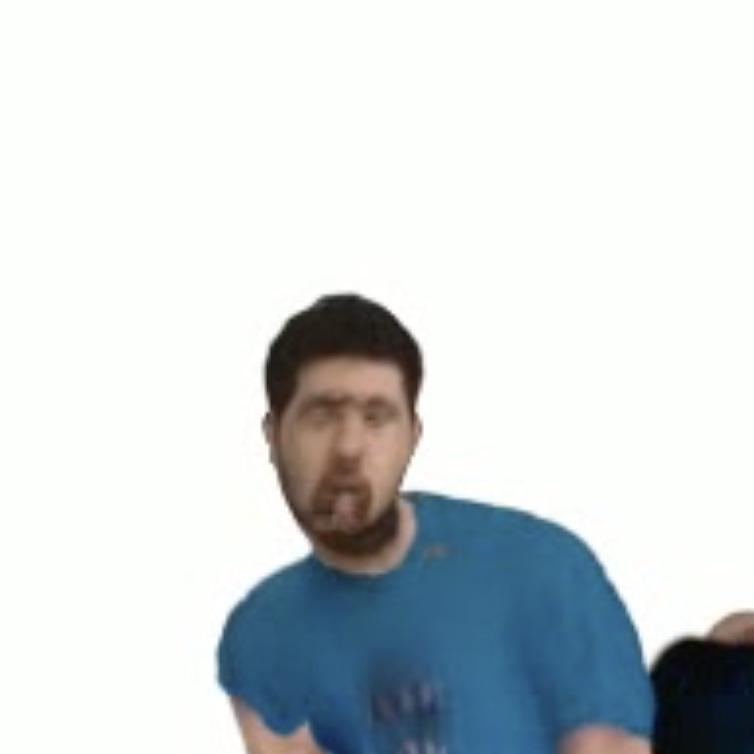}} &
        \fbox{\includegraphics[width=0.15\textwidth]{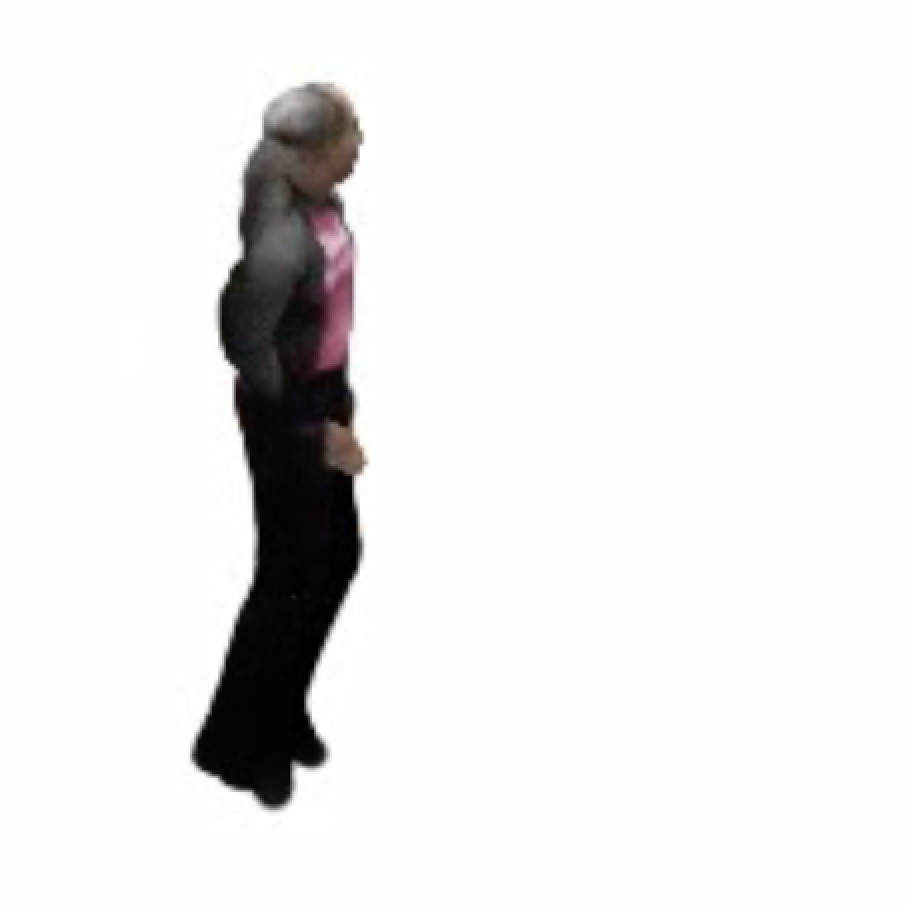}} &
        \fbox{\includegraphics[width=0.15\textwidth]{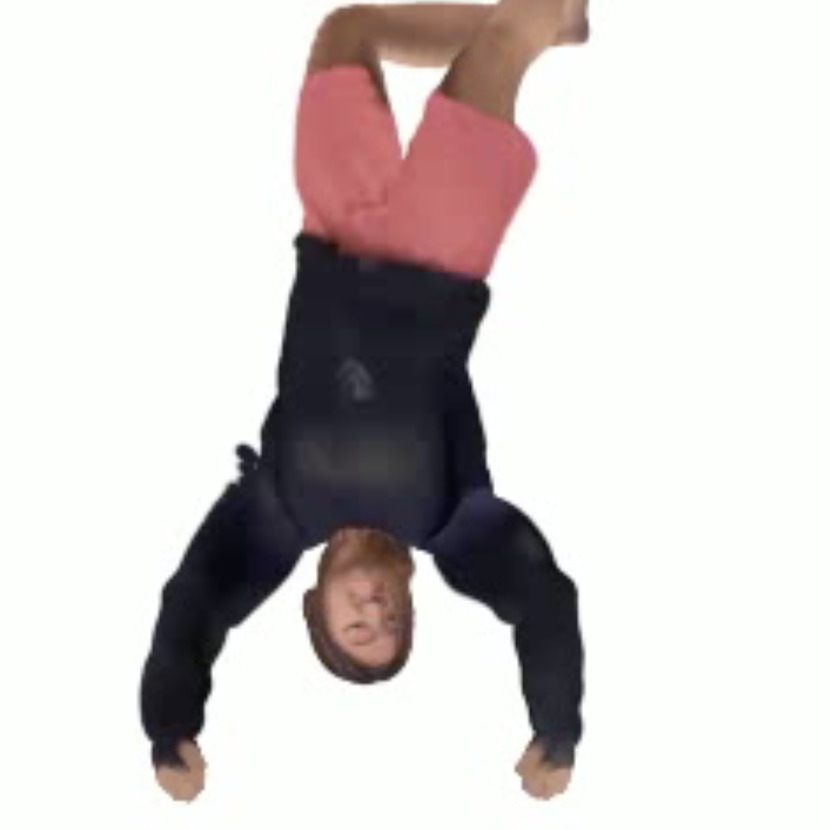}} \\

        \fbox{\includegraphics[width=0.15\textwidth]{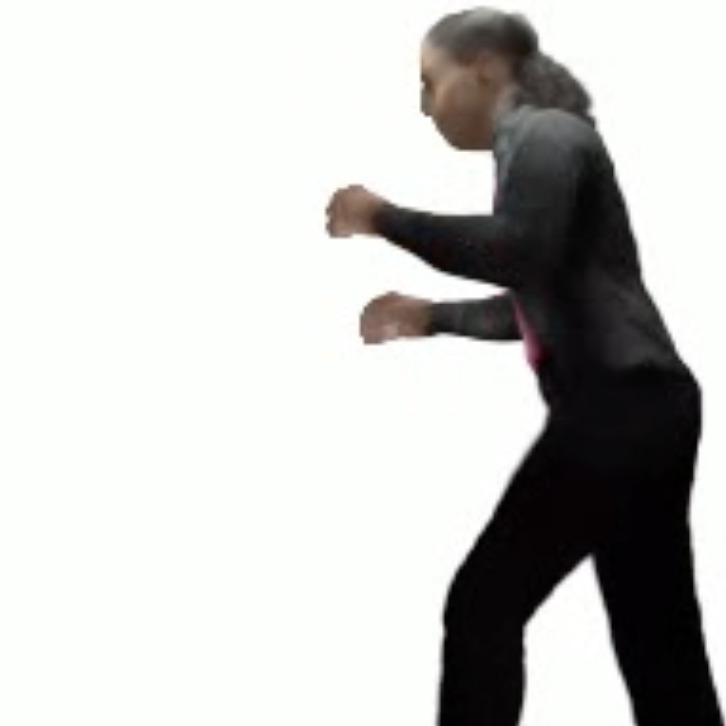}} &
        \fbox{\includegraphics[width=0.15\textwidth]{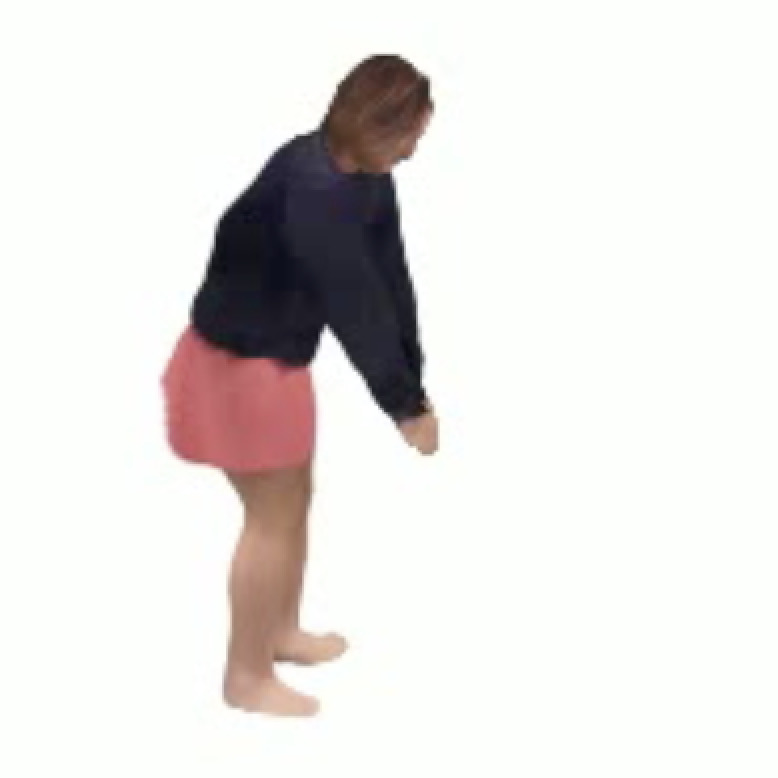}} &
        \fbox{\includegraphics[width=0.15\textwidth]{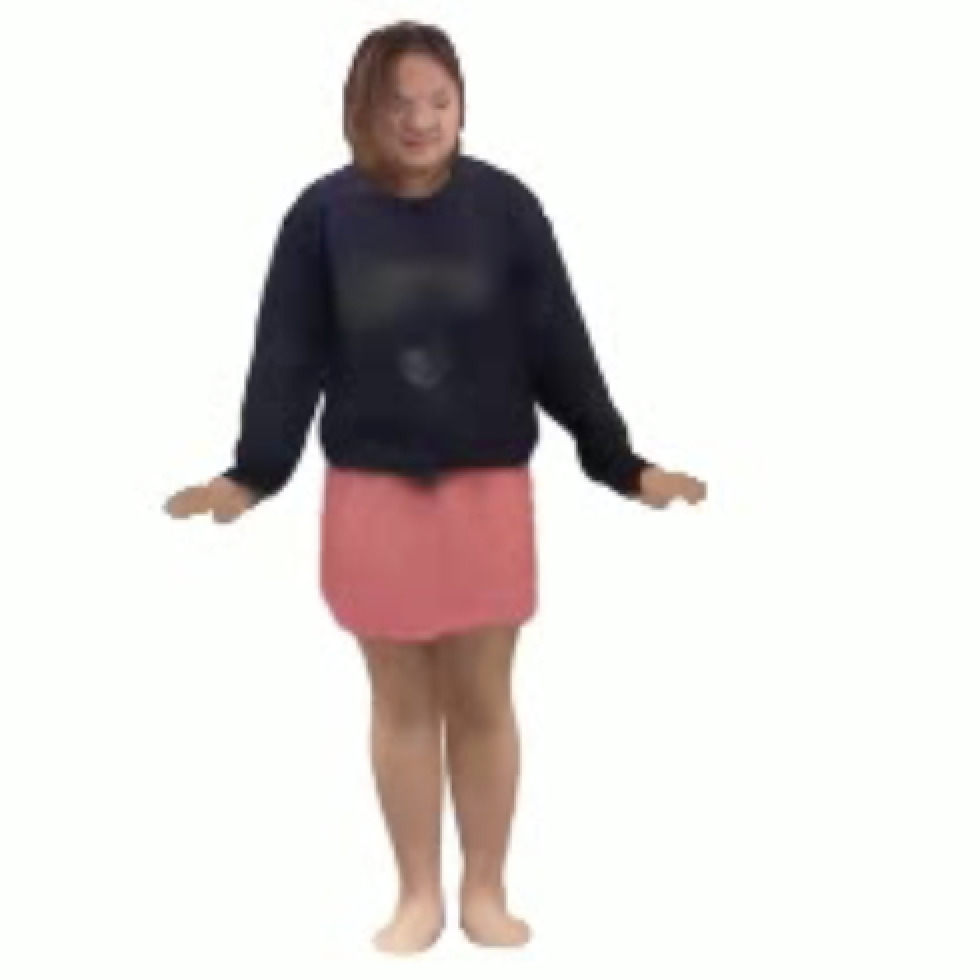}} &
        \fbox{\includegraphics[width=0.15\textwidth]{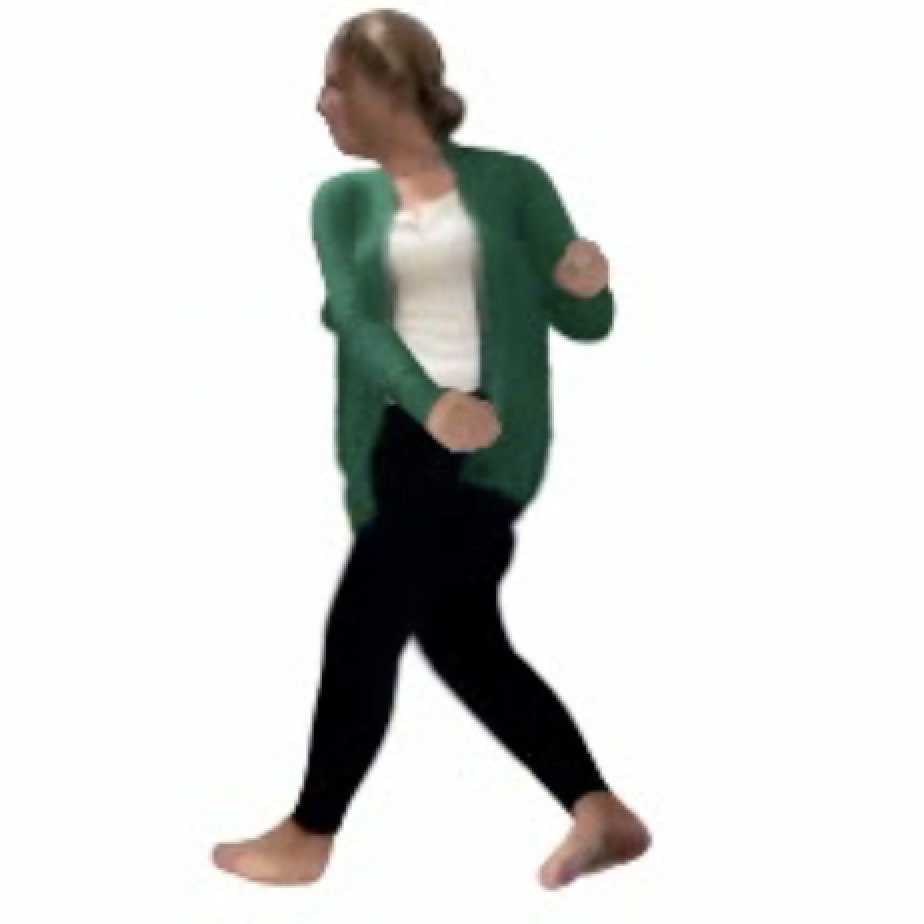}} &
        \fbox{\includegraphics[width=0.15\textwidth]{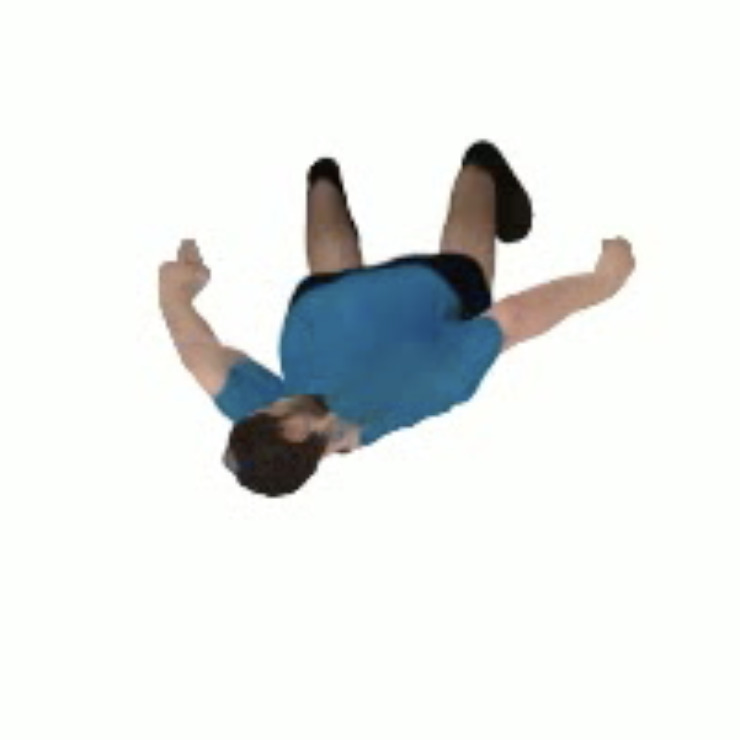}} &
        \fbox{\includegraphics[width=0.15\textwidth]{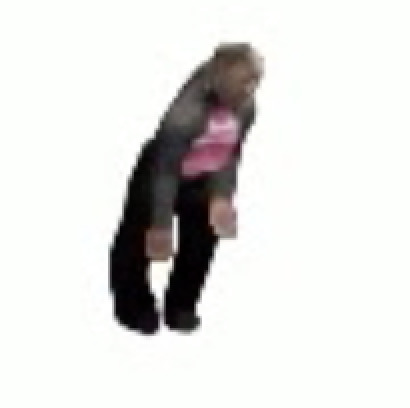}} \\

        \fbox{\includegraphics[width=0.15\textwidth]{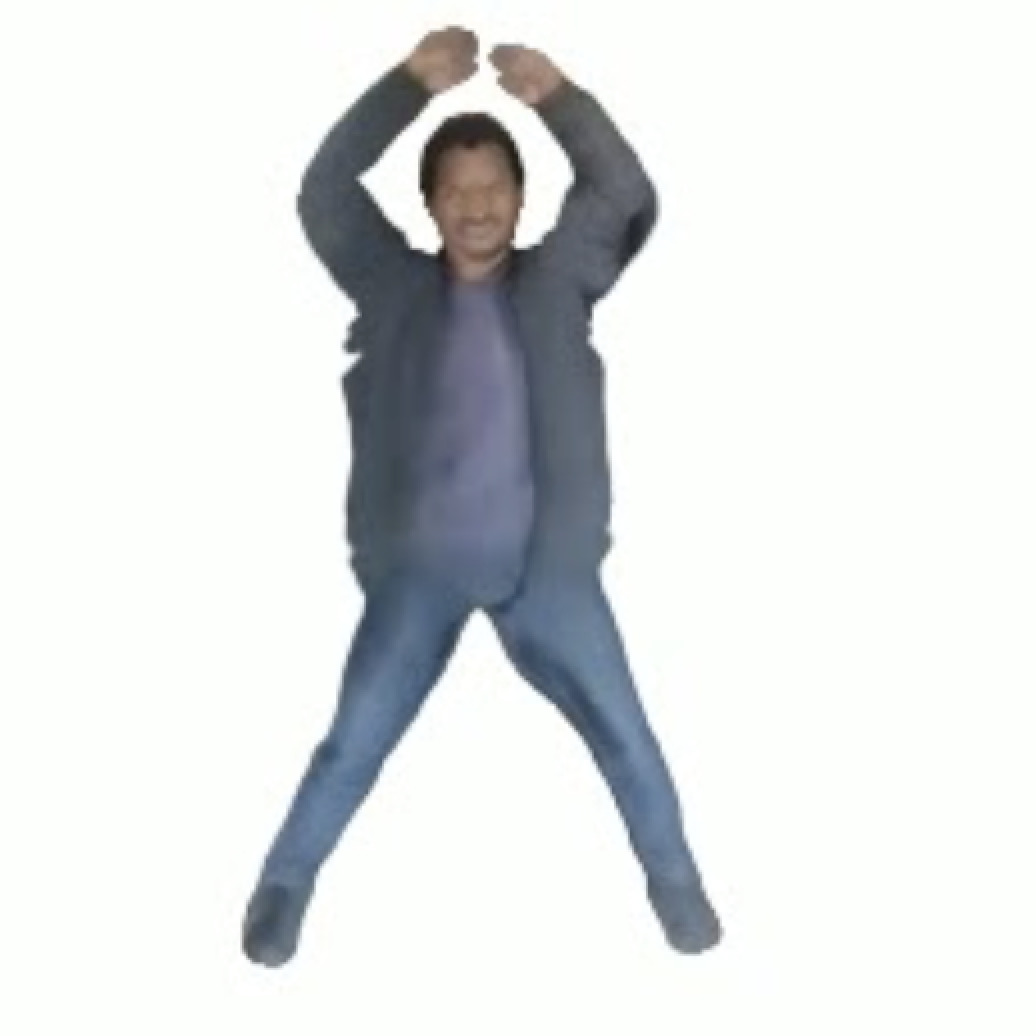}} &
        \fbox{\includegraphics[width=0.15\textwidth]{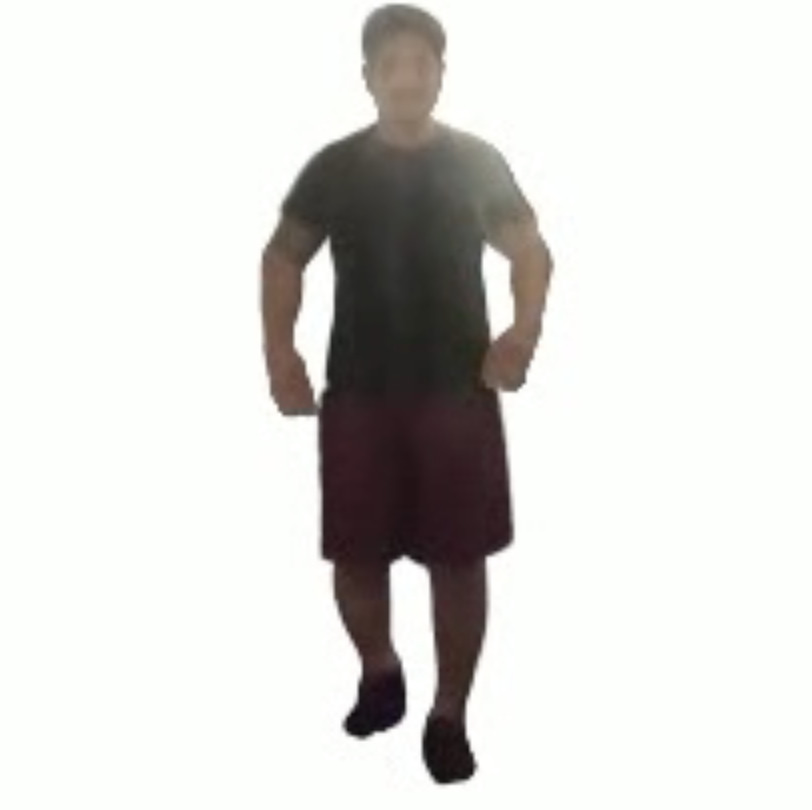}} &
        \fbox{\includegraphics[width=0.15\textwidth]{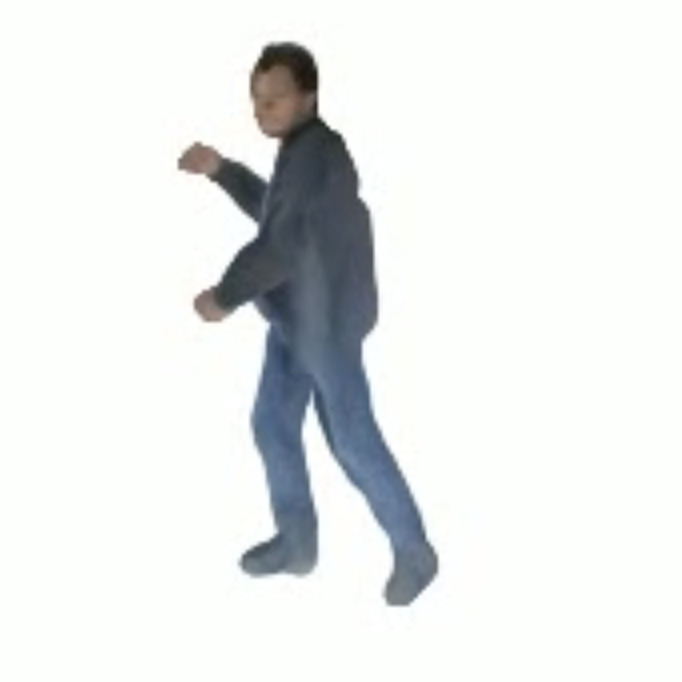}} &
        \fbox{\includegraphics[width=0.15\textwidth]{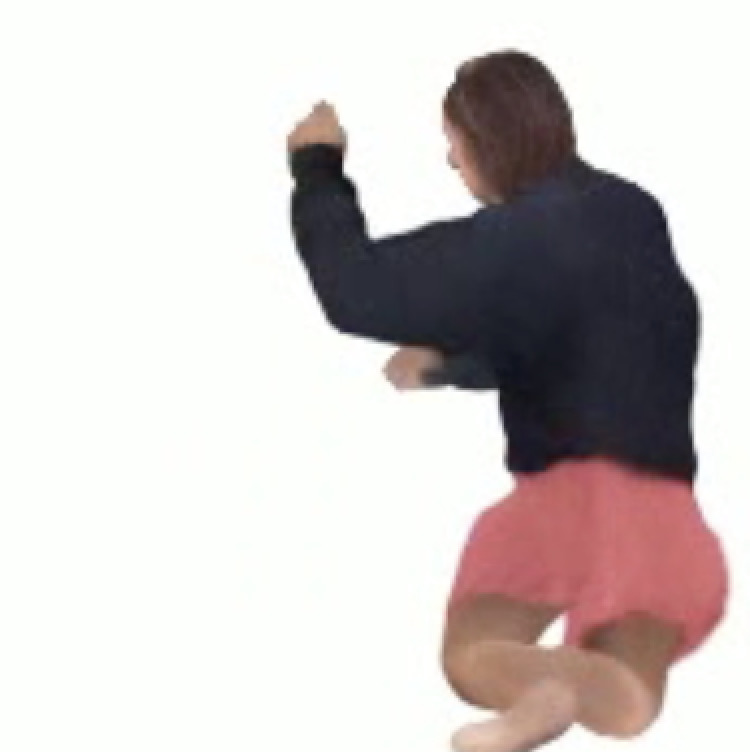}} &
        \fbox{\includegraphics[width=0.15\textwidth]{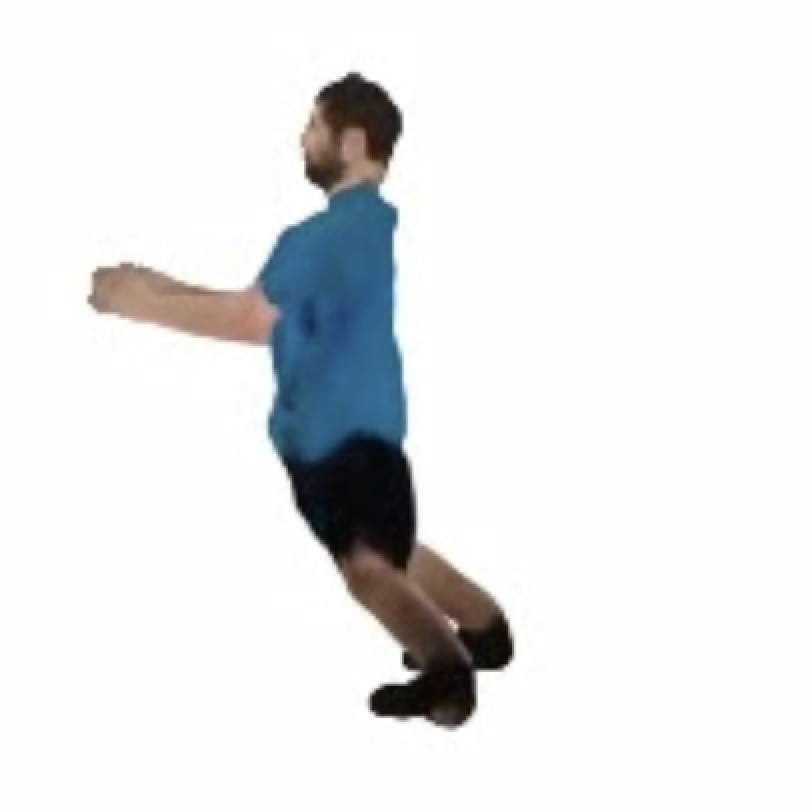}} &
        \fbox{\includegraphics[width=0.15\textwidth]{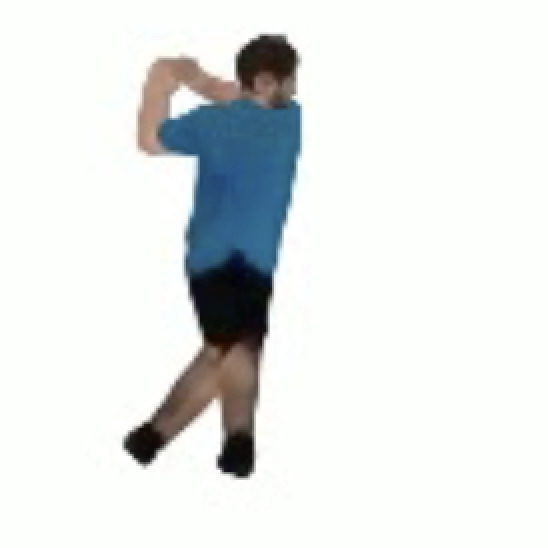}} \\

        \multicolumn{3}{c}{\small Consistent and photorealistic frames} \hspace{26pt} &
        \multicolumn{3}{c}{\small Inconsistent or non-photorealistic frames} 
    \end{tabular}
    \caption{Representative examples of video frames generated by \textit{ExAvatar}.}
    \label{fig:examples-exavatar}
\end{figure*}

\textbf{ExAvatar. } This method represents the novel human identity as a video (in our case, sourced from the RANDOM People dataset). It generates an expressive whole-body 3D Gaussian avatar and animates it following the pose in the reference action video (source). The official implementation of \textit{ExAvatar}\footnote{\url{https://github.com/mks0601/ExAvatar_RELEASE}} was used. Examples of representative video frames generated by \textit{ExAvatar} are shown in Figure~\ref{fig:examples-exavatar}.

%\begin{figure}[t]
%    \centering
%    \begin{tabular}{c@{\hskip 4pt}c@{\hskip 4pt}c}
%        \includegraphics[width=0.31\linewidth]{fig/fig-2/frame-ucf-squat.jpg} &
%        \includegraphics[width=0.31\linewidth]{fig/fig-2/frame-ucf-jack.jpg} &
%        \includegraphics[width=0.31\linewidth]{fig/fig-2/frame-ucf-golf.jpg} \\
%        \small (a) Bodyweight squats & \small (b) Jumping jack & \small (c) Golf swing
%    \end{tabular}
%    \caption{Representative examples of frames from UCF101, cropped to fit the layout.}
%    \label{fig:ucf101-videos}
%\end{figure}

\section{Survey Methodology}
\label{sec:survey-methodology}

We recruited $16$ participants through snowball sampling to complete a survey with $66$ questions hosted on Qualtrics. This survey comprised two tasks with quantitative questions ($2 \times 30$ questions in total) and one task with qualitative questions ($6$ questions). All videos used in the survey were chosen randomly; each participant saw the same videos in the same order. There is an equal number of videos generated by each of the pose transfer methods: \textit{AnimateAnyone}, \textit{MagicAnimate}, and \textit{ExAvatar}. The wording of participant instructions and questions are presented in Appendix~\ref{app:survey-instructions-and-questions}.

\textbf{Task 1: Generated Action Evaluation. } This section contained $30$ questions that presented participants with a pose-transferred video and asked them to select one of $20$ action classes (matched to the subset of $20$ classes from UCF101, see Section~\ref{sec:methods}) that best describes the action performed in the generated video.

\textbf{Task 2: Transfer Control Evaluation. } This section contained $30$ questions that presented participants with two videos: a pose-transferred video and its source video. The questions then ask the participants to indicate if the action in the pose-transferred video is consistent with the source video; the participants could label the pair as 'consistent', 'somewhat consistent', or 'inconsistent'.

\textbf{Task 3: Qualitative Evaluation. } This section contained $6$ questions that presented participants with a single video and asked them a series of free-form questions about its quality.

\section{Results}

This section presents the results of our survey, conducted over two weeks in November 2024. To enhance the reproducibility of our work, we open-source the response data in full at \url{https://github.com/matyasbohacek/pose-transfer-human-motion}.

\textbf{Task 1: Generated Action Evaluation. } Overall, the correct action\footnote{The correct action is the action shown in the reference (source) video.} (out of $20$ action classes) was selected $42.92 \%$ of the time. When broken down to individual pose transfer methods, the participants selected the correct action the least for \textit{MagicAnimate} videos, $13.12 \%$ of the time. For \textit{AnimateAnyone}, the participants selected the correct action $47.50 \%$ of the time, and for \textit{ExAvatar}, they did so $68.12 \%$ of the time.

\begin{table}[t]
    \centering
    \small % Reducing the font size
    \begin{tabularx}{\linewidth}{lXXX} 
        \toprule
        & \raisebox{-2.85ex}{\textbf{Consistent}} 
        & \textbf{Partially Consistent}
        & \raisebox{-2.85ex}{\textbf{Inconsistent}} \\
        \midrule
        \textbf{AnimateAnyone} & 27.50 & 25.62 & 46.88 \\
        \textbf{MagicAnimate} & 26.88 & 31.25 & 41.88 \\
        \textbf{ExAvatar} & 55.00 & 18.12 & 38.54 \\
        \midrule
        \textbf{Overall} & 36.46 & 25.00 & 42.43 \\
        \bottomrule
    \end{tabularx}
    \caption{Results of the Transfer Control Evaluation (Task 1), reported as percentages per each pose-transfer method. The overall results is an average of the above.}
    \label{tab:eval-one}
\end{table}

\textbf{Task 2: Transfer Control Evaluation. } The results are shown in Table~\ref{tab:eval-one}. Overall, the video pairs were labeled consistent $36.46 \%$ of the time, partially consistent $25.00 \%$ of the time, and inconsistent $42.43 \%$ of the time. When broken down to individual pose transfer methods, \textit{MagicAnimate} videos were deemed the least consistent---participants labeled them consistent only $26.88 \%$ of the time, partially consistent $31.25 \%$ of the time, and inconsistent $41.88 \%$ of the time. The \textit{AnimateAnyone} videos came slightly above: participants labeled them consistent $27.50 \%$ of the time, partially consistent $25.62 \%$ of the time, and inconsistent $46.88 \%$ of the time. Finally, the \textit{ExAvatar} videos come on top: participants labeled them as consistent $55.00 \%$ of the time, partially consistent $18.12 \%$ of the time, and as inconsistent $38.54 \%$ of the time.

\textbf{Task 3: Qualitative Evaluation. } Participants commonly noted that the videos across all pose transfer methods lacked photorealism and consistency. For example, P14 reacted to Q3.1 with "\textit{not photorealistic, [I] dont know what he is doing}", and P12 said in reaction to Q3.4 "\textit{that is not photorealistic, it [is] not very consistent}". Interestingly, some participants said that, despite these shortcomings, the action was recognizable to them. For example, in response to Q3.1, P13 shared that "\textit{The person is not photorealistic. Action seems to be jumping jack but might be something else.}" In response to Q3.2, P11 said that "\textit{His movements are so small and slow that I can't tell what he's doing. It looks like Tai Chi, but if I didn't know it was supposed to be Tai Chi, I wouldn't have guessed.}" 

While the responses were mostly similar across the pose transfer methods, one difference emerged between the diffusion-based methods, \textit{AnimateAnyone} and \textit{MagicAnimate}, and the splatting-based \textit{ExAvatar}: the videos generated by the diffusion-based methods were often found to have structural deficiencies in the generated human bodies. For example, P2 pointed out that the videos evaluated in Q3.1 and Q3.2 show a "\textit{disorted face}" and "\textit{disorted hands}". In reaction to Q3.3, P8 said that "\textit{another person appears behind (...) and it's weird.}" For Q3.4, they added that "\textit{The hair changes and she's missing one foot.}" The \textit{ExAvatar} videos, on the other hand, have not been found to evince similar artifacts by the participants.

\section{Discussion}

The results of our survey reveal that, in a controlled environment of $20$ distinct OOD human actions, state-of-the-art pose transfer methods largely fail to generate consistent and photorealistic videos that convey understandable actions. 

When asked to select which action (out of a list of $20$ actions) is most likely performed in the \textit{MagicAnimate} videos, the participants selected the correct action only $13.12 \%$ of the time. This is particularly striking given that, had the participants selected the action randomly, they would have gotten the correct action $5 \%$ of the time. The other diffusion-based method we evaluated, \textit{AnimateAnyone}, seemed to convey the semantics of the action better, as the participants selected the correct action $47.50 \%$ of the time. Interestingly, despite this stark difference in the ability to select the correct action (Task 1), the participants perceived the consistency of the source video and the generated video (Task 2) for these two methods similarly. In particular, the \textit{AnimateAnyone} and \textit{MagicAnimate} videos were deemed consistent $27.50 \%$ and $26.88 \%$ of the time, partially consistent $25.62 \%$ and $31.25 \%$ of time, and inconsistent $46.88 \%$ and $41.88 \%$ of time, respectively. We speculate that, despite the \textit{AnimateAnyone} videos conveying the action semantics better, the presence of undesired artifacts still leads participants to fixate on the poor generation quality. 

The \textit{ExAvatar} method outperformed its counterparts in our study, with participants selecting the correct action $68.12\%$ of the time and labeling $55.00\%$ of the videos as consistent. However, a significant amount of the videos, $38.54 \%$, were still labeled as inconsistent, on par with \textit{AnimateAnyone} and \textit{MagicAnimate}, whose videos were labeled inconsistent $46.88 \%$ and $41.88 \%$ of the time, respectively. While this splatting-based method mitigates many rendering issues observed in diffusion-based methods---such as implausible body poses and inconsistent identities---its photorealism and fidelity still require improvement for the actions to be more recognizable.

Taken together, these results suggest that the splatting-based \textit{ExAvatar} produces more photorealistic, consistent, and controllable human motion than the diffusion-based \textit{AnimateAnyone} and \textit{MagicAnimate}. This, however, comes at the cost of a more demanding input (novel human identity reference): while \textit{AnimateAnyone} and \textit{MagicAnimate} require only a static image, \textit{ExAvatar} requires a video in which the protagonist performs a sequence of movements described in~\cite{moon2024expressive}. 

In either case, there remains much work to make pose transfer consistent and photorealistic, while minimizing the input demands. We stipulate that combining 3D representations (and, more broadly, traditional computer graphics methods) with recent generative AI methods holds much promise in addressing these issues. In particular, such methods can leverage the high controllability and consistency of identities from the 3D methods with high efficiency and low input requirements of generative AI methods.

The surprisingly poor performance of the evaluated pose transfer methods on OOD data with human feedback highlights a gap in the literature. Current standardized benchmarks may not adequately assess the ability of these methods to generalize outside of their training domain, particularly for novel activities and human identities. Developing more challenging benchmarks that test generalization capabilities and employing human evaluations to assess action semantics in generated videos could address this gap and accelerate progress in pose transfer methods.

\section{Conclusion}

State-of-the-art pose transfer methods continue to face challenges in generating consistent and photorealistic videos that accurately preserve the motion semantics of source videos. Among the evaluated methods, we found videos generated by the splatting-based \textit{ExAvatar} more consistent and photorealistic than videos generated by the diffusion-based \textit{AnimateAnyone} and \textit{MagicAnimate}.

Our participant survey highlights several key areas for future improvement in pose transfer methods: (1) enhancing photorealism, (2) maintaining greater consistency with source videos, and (3) improving the fidelity of novel human identities. Moreover, given the saturation of existing benchmarks, we underscore the need for a new pose transfer benchmark that would evaluate the generalization abilities of pose transfer methods moving forward.

% \section*{Acknowledgements}
% Anonymized
% We would like to extend our thanks to TBD.

{\small
\bibliographystyle{ieee_fullname}
\bibliography{egbib}

\begin{thebibliography}{10}\itemsep=-1pt

\bibitem{chen2023open}
Junyang Chen, Xiaoyu Xian, Zhijing Yang, Tianshui Chen, Yongyi Lu, Yukai Shi, Jinshan Pan, and Liang Lin.
\newblock Open-world pose transfer via sequential test-time adaption.
\newblock {\em arXiv preprint arXiv:2303.10945}, 2023.

\bibitem{chen2022transfer}
Shuhong Chen and Matthias Zwicker.
\newblock Transfer learning for pose estimation of illustrated characters.
\newblock In {\em Proceedings of the IEEE/CVF winter conference on applications of computer vision}, pages 793--802, 2022.

\bibitem{dubey2023comprehensive}
Shradha Dubey and Manish Dixit.
\newblock A comprehensive survey on human pose estimation approaches.
\newblock {\em Multimedia Systems}, 29(1):167--195, 2023.

\bibitem{ekambaram2024real}
Dilliraj Ekambaram and Vijayakumar Ponnusamy.
\newblock Real-time ai-assisted visual exercise pose correctness during rehabilitation training for musculoskeletal disorder.
\newblock {\em Journal of Real-Time Image Processing}, 21(1):2, 2024.

\bibitem{ghodhbani2022you}
Hajer Ghodhbani, Mohamed Neji, Imran Razzak, and Adel~M Alimi.
\newblock You can try without visiting: a comprehensive survey on virtually try-on outfits.
\newblock {\em Multimedia Tools and Applications}, 81(14):19967--19998, 2022.

\bibitem{güler2018denseposedensehumanpose}
Rıza~Alp Güler, Natalia Neverova, and Iasonas Kokkinos.
\newblock Densepose: Dense human pose estimation in the wild, 2018.

\bibitem{haleem2022human}
Mohammad~Baqer Haleem and Israa~Hadi Ali.
\newblock Human pose transfer based on deep learning models: A survey.
\newblock In {\em 2022 3rd Information Technology To Enhance e-learning and Other Application (IT-ELA)}, pages 163--169. IEEE, 2022.

\bibitem{hinton2008visualizing}
G Hinton and L Van Der~Maaten.
\newblock Visualizing data using t-sne journal of machine learning research.
\newblock {\em Journal of Machine Learning Research}, 9:2579--2605, 2008.

\bibitem{hu2024animateanyoneconsistentcontrollable}
Li Hu, Xin Gao, Peng Zhang, Ke Sun, Bang Zhang, and Liefeng Bo.
\newblock Animate anyone: Consistent and controllable image-to-video synthesis for character animation, 2024.

\bibitem{jafarian2021learning}
Yasamin Jafarian and Hyun~Soo Park.
\newblock Learning high fidelity depths of dressed humans by watching social media dance videos.
\newblock In {\em Proceedings of the IEEE/CVF Conference on Computer Vision and Pattern Recognition}, pages 12753--12762, 2021.

\bibitem{jiang2024cinematic}
Xuekun Jiang, Anyi Rao, Jingbo Wang, Dahua Lin, and Bo Dai.
\newblock Cinematic behavior transfer via nerf-based differentiable filming.
\newblock In {\em Proceedings of the IEEE/CVF Conference on Computer Vision and Pattern Recognition}, pages 6723--6732, 2024.

\bibitem{kappel2021high}
Moritz Kappel, Vladislav Golyanik, Mohamed Elgharib, Jann-Ole Henningson, Hans-Peter Seidel, Susana Castillo, Christian Theobalt, and Marcus Magnor.
\newblock High-fidelity neural human motion transfer from monocular video.
\newblock In {\em Proceedings of the IEEE/CVF conference on computer vision and pattern recognition}, pages 1541--1550, 2021.

\bibitem{karras2023dreamposefashionimagetovideosynthesis}
Johanna Karras, Aleksander Holynski, Ting-Chun Wang, and Ira Kemelmacher-Shlizerman.
\newblock Dreampose: Fashion image-to-video synthesis via stable diffusion, 2023.

\bibitem{khan2024human}
Muhammad Saif~Ullah Khan, Muhammad~Ferjad Naeem, Federico Tombari, Luc Van~Gool, Didier Stricker, and Muhammad~Zeshan Afzal.
\newblock Human pose descriptions and subject-focused attention for improved zero-shot transfer in human-centric classification tasks.
\newblock {\em arXiv preprint arXiv:2403.06904}, 2024.

\bibitem{kumar2024pose}
Lalit Kumar and Dushyant~Kumar Singh.
\newblock Pose image generation for video content creation using controlled human pose image generation gan.
\newblock {\em Multimedia Tools and Applications}, 83(20):59335--59354, 2024.

\bibitem{kumar2022human}
Pranjal Kumar, Siddhartha Chauhan, and Lalit~Kumar Awasthi.
\newblock Human pose estimation using deep learning: review, methodologies, progress and future research directions.
\newblock {\em International Journal of Multimedia Information Retrieval}, 11(4):489--521, 2022.

\bibitem{li2023collecting}
Nannan Li, Kevin~J Shih, and Bryan~A Plummer.
\newblock Collecting the puzzle pieces: Disentangled self-driven human pose transfer by permuting textures.
\newblock In {\em Proceedings of the IEEE/CVF International Conference on Computer Vision}, pages 7126--7137, 2023.

\bibitem{LIU2021107024}
Meichen Liu, Kejun Wang, Ruihang Ji, Shuzhi~Sam Ge, and Jing Chen.
\newblock Pose transfer generation with semantic parsing attention network for person re-identification.
\newblock {\em Knowledge-Based Systems}, 223:107024, 2021.

\bibitem{liu2023poseexaminer}
Qihao Liu, Adam Kortylewski, and Alan~L Yuille.
\newblock {PoseExaminer}: Automated testing of out-of-distribution robustness in human pose and shape estimation.
\newblock In {\em Proceedings of the IEEE/CVF Conference on Computer Vision and Pattern Recognition}, pages 672--681, 2023.

\bibitem{liu2019liquid}
Wen Liu, Zhixin Piao, Jie Min, Wenhan Luo, Lin Ma, and Shenghua Gao.
\newblock Liquid warping {GAN}: A unified framework for human motion imitation, appearance transfer and novel view synthesis.
\newblock In {\em Proceedings of the IEEE/CVF international conference on computer vision}, pages 5904--5913, 2019.

\bibitem{ma2017pose}
Liqian Ma, Xu Jia, Qianru Sun, Bernt Schiele, Tinne Tuytelaars, and Luc Van~Gool.
\newblock Pose guided person image generation.
\newblock {\em Advances in neural information processing systems}, 30, 2017.

\bibitem{moon2024expressive}
Gyeongsik Moon, Takaaki Shiratori, and Shunsuke Saito.
\newblock Expressive whole-body 3d gaussian avatar.
\newblock {\em arXiv preprint arXiv:2407.21686}, 2024.

\bibitem{pedregosa2011scikit}
Fabian Pedregosa, Ga{\"e}l Varoquaux, Alexandre Gramfort, Vincent Michel, Bertrand Thirion, Olivier Grisel, Mathieu Blondel, Peter Prettenhofer, Ron Weiss, Vincent Dubourg, et~al.
\newblock Scikit-learn: Machine learning in python.
\newblock {\em the Journal of machine Learning research}, 12:2825--2830, 2011.

\bibitem{podell2023sdxl}
Dustin Podell, Zion English, Kyle Lacey, Andreas Blattmann, Tim Dockhorn, Jonas M{\"u}ller, Joe Penna, and Robin Rombach.
\newblock Sdxl: Improving latent diffusion models for high-resolution image synthesis.
\newblock {\em arXiv preprint arXiv:2307.01952}, 2023.

\bibitem{radford2021learning}
Alec Radford, Jong~Wook Kim, Chris Hallacy, Aditya Ramesh, Gabriel Goh, Sandhini Agarwal, Girish Sastry, Amanda Askell, Pamela Mishkin, Jack Clark, et~al.
\newblock Learning transferable visual models from natural language supervision.
\newblock In {\em International conference on machine learning}, pages 8748--8763. PMLR, 2021.

\bibitem{radford2021learningtransferablevisualmodels}
Alec Radford, Jong~Wook Kim, Chris Hallacy, Aditya Ramesh, Gabriel Goh, Sandhini Agarwal, Girish Sastry, Amanda Askell, Pamela Mishkin, Jack Clark, Gretchen Krueger, and Ilya Sutskever.
\newblock Learning transferable visual models from natural language supervision, 2021.

\bibitem{ren2020human}
Jian Ren, Menglei Chai, Sergey Tulyakov, Chen Fang, Xiaohui Shen, and Jianchao Yang.
\newblock Human motion transfer from poses in the wild.
\newblock In {\em Computer Vision--ECCV 2020 Workshops: Glasgow, UK, August 23--28, 2020, Proceedings, Part III 16}, pages 262--279. Springer, 2020.

\bibitem{ren2020humanmotiontransferposes}
Jian Ren, Menglei Chai, Sergey Tulyakov, Chen Fang, Xiaohui Shen, and Jianchao Yang.
\newblock Human motion transfer from poses in the wild, 2020.

\bibitem{roy2022multiscaleattentionguidedpose}
Prasun Roy, Saumik Bhattacharya, Subhankar Ghosh, and Umapada Pal.
\newblock Multi-scale attention guided pose transfer, 2022.

\bibitem{shen2023x}
Kaiyue Shen, Chen Guo, Manuel Kaufmann, Juan~Jose Zarate, Julien Valentin, Jie Song, and Otmar Hilliges.
\newblock X-avatar: Expressive human avatars.
\newblock In {\em Proceedings of the IEEE/CVF Conference on Computer Vision and Pattern Recognition}, pages 16911--16921, 2023.

\bibitem{shukla2022vl4pose}
Megh Shukla, Roshan Roy, Pankaj Singh, Shuaib Ahmed, and Alexandre Alahi.
\newblock {VL4Pose}: Active learning through out-of-distribution detection for pose estimation.
\newblock {\em arXiv preprint arXiv:2210.06028}, 2022.

\bibitem{siarohin2018deformable}
Aliaksandr Siarohin, Enver Sangineto, St{\'e}phane Lathuiliere, and Nicu Sebe.
\newblock Deformable {GANs} for pose-based human image generation.
\newblock In {\em Proceedings of the IEEE conference on computer vision and pattern recognition}, pages 3408--3416, 2018.

\bibitem{siarohin2021motion}
Aliaksandr Siarohin, Oliver~J Woodford, Jian Ren, Menglei Chai, and Sergey Tulyakov.
\newblock Motion representations for articulated animation.
\newblock In {\em Proceedings of the IEEE/CVF Conference on Computer Vision and Pattern Recognition}, pages 13653--13662, 2021.

\bibitem{song2023image}
Dan Song, Xuanpu Zhang, Juan Zhou, Weizhi Nie, Ruofeng Tong, Mohan Kankanhalli, and An-An Liu.
\newblock Image-based virtual try-on: A survey.
\newblock {\em arXiv preprint arXiv:2311.04811}, 2023.

\bibitem{soomro2012ucf101dataset101human}
Khurram Soomro, Amir~Roshan Zamir, and Mubarak Shah.
\newblock Ucf101: A dataset of 101 human actions classes from videos in the wild, 2012.

\bibitem{wang2023edittemporalconsistentvideosimage}
Yuanzhi Wang, Yong Li, Xiaoya Zhang, Xin Liu, Anbo Dai, Antoni~B. Chan, and Zhen Cui.
\newblock Edit temporal-consistent videos with image diffusion model, 2023.

\bibitem{wu2023human}
Kun Wu, Chengxiang Yin, Zhengping Che, Bo Jiang, Jian Tang, Zheng Guan, and Gangyi Ding.
\newblock Human pose transfer with augmented disentangled feature consistency.
\newblock {\em ACM Transactions on Intelligent Systems and Technology}, 15(1):1--22, 2023.

\bibitem{xu2023magicanimatetemporallyconsistenthuman}
Zhongcong Xu, Jianfeng Zhang, Jun~Hao Liew, Hanshu Yan, Jia-Wei Liu, Chenxu Zhang, Jiashi Feng, and Mike~Zheng Shou.
\newblock Magicanimate: Temporally consistent human image animation using diffusion model, 2023.

\bibitem{yu2023bidirectionally}
Wing-Yin Yu, Lai-Man Po, Ray~CC Cheung, Yuzhi Zhao, Yu Xue, and Kun Li.
\newblock Bidirectionally deformable motion modulation for video-based human pose transfer.
\newblock In {\em Proceedings of the IEEE/CVF International Conference on Computer Vision}, pages 7502--7512, 2023.

\bibitem{zablotskaia2019dwnet}
Polina Zablotskaia, Aliaksandr Siarohin, Bo Zhao, and Leonid Sigal.
\newblock Dwnet: Dense warp-based network for pose-guided human video generation.
\newblock {\em arXiv preprint arXiv:1910.09139}, 2019.

\bibitem{zheng2023deep}
Ce Zheng, Wenhan Wu, Chen Chen, Taojiannan Yang, Sijie Zhu, Ju Shen, Nasser Kehtarnavaz, and Mubarak Shah.
\newblock Deep learning-based human pose estimation: A survey.
\newblock {\em ACM Computing Surveys}, 56(1):1--37, 2023.

\bibitem{zhu2023human}
Wentao Zhu, Xiaoxuan Ma, Dongwoo Ro, Hai Ci, Jinlu Zhang, Jiaxin Shi, Feng Gao, Qi Tian, and Yizhou Wang.
\newblock Human motion generation: A survey.
\newblock {\em IEEE Transactions on Pattern Analysis and Machine Intelligence}, 2023.

\bibitem{zhu2019progressive}
Zhen Zhu, Tengteng Huang, Baoguang Shi, Miao Yu, Bofei Wang, and Xiang Bai.
\newblock Progressive pose attention transfer for person image generation.
\newblock In {\em Proceedings of the IEEE/CVF conference on computer vision and pattern recognition}, pages 2347--2356, 2019.

\end{thebibliography}
}

\newpage

~

\newpage

\appendix

%
%
% \pagenumbering{gobble}
% \fancyhead{}
% \renewcommand{\headrulewidth}{0pt}
%
%

%
\section{Selected Action Classes}
\label{app:selected-action-classes}

The following classes from UCF101 were used for the reference action videos:

\vspace{0.10cm}

\noindent
\begin{enumerate}
    \item Apply Lipstick
    \item Archery
    \item Body Weight Squats
    \item Clean And Jerk
    \item Golf Swing
    \item Hammer Throw
    \item Hammering
    \item Hand Stand Pushups
    \item Hula Hoop
    \item Jumping Jack
    \item Lunges
    \item Mopping Floor
    \item Pull Ups
    \item Shot Put
    \item Soccer Juggling
    \item Table Tennis Shot
    \item Tai Chi
    \item Tennis Swing
    \item Throw Discus
    \item Writing On Board
\end{enumerate}

\section{Survey Instructions and Questions}
\label{app:survey-instructions-and-questions}

The following instructions and questions were presented to participants in Qualtrics:

\noindent\rule{\linewidth}{0.4pt}
\textbf{Introduction. } This survey examines the quality of generative AI models for human action generation. You will be presented with questions about the quality and semantics of human action videos. Please play each video carefully before you answer. Videos may be rewatched without restriction.

\noindent\rule{\linewidth}{0.4pt}
\textbf{Section 1. } The following questions will ask you to classify actions in short videos out of a set of 20 activity names. Choose the activity name that you think best describes the shown action. At each step, play the video and then answer the question.

\noindent\rule{\linewidth}{0.4pt}
\textbf{Q1.1-Q1.20. } Which of the following activity names best describes the action shown in the video above?

\noindent\rule{\linewidth}{0.4pt}
\textbf{Section 2. } The following questions will ask you to decide if pairs of short videos show consistent activities. As such, the precise body movement must not be synchronized; it is only the overall activity name that must be the same. For example, if the videos will both show a person sitting down, that would be a constant activity. On the contrary, if one video shows a person sitting down and the other shows a person riding a horse, those would be inconsistent activities. At each step, play both videos and then answer the question.

\noindent\rule{\linewidth}{0.4pt}
\textbf{Q2.1-Q2.20. } Are the activities shown in the two videos above consistent?

\noindent\rule{\linewidth}{0.4pt}
\textbf{Section 3. } This last section will ask you to provide free-form evaluation of the quality of five short videos. Comment on any—or all—of the following questions. At each step, play the video and then answer the question. \textbf{1.} Is the person shown in the video photorealistic—that is, is their body rendered properly and is the identity consistent through the video? \textbf{2.} Is the person shown performing a clearly identifiable action? \textbf{3.} Do you have any other comments that come to mind? 

% TODO MB

\section{Responses to the Qualitative Evaluation}
\label{app:responses-to-the-qualitative-evaluation}

The complete set of free-form participant responses to Task 3: Qualitative Evaluation, which includes Q3.1, Q3.2, Q3.3, Q3.4, Q3.5, and Q3.6, is presented in the following three tables, organized by the model used to generate the evaluated videos. Table~\ref{tab:qual-results-1} presents the results for \textit{AnimateAnyone}; Table~\ref{tab:qual-results-2} presents the results for \textit{MagicAnimate}; and Table~\ref{tab:qual-results-3} presents the results for \textit{ExAvatar}. This data is also available in CSV format at \url{https://github.com/matyasbohacek/pose-transfer-human-motion}.

\onecolumn
\newpage

\newpage

\begin{table*}[h]
\centering
\begin{tabular}{|l|p{0.4\textwidth}|p{0.4\textwidth}|}
\hline
\textbf{ID} & \textbf{Q3.1} & \textbf{Q3.2} \\ \hline
1 & The person shown on the video is distorted, he is not photorealistic. It is rendered decently. By the body shape you can tell that it is male. His actions looks like he is performing jumping jacks. & On this video, there is a person of a male gender, he looks much more identifiable than the person from previous video, render in this video is somewhat good. It looks like he is performing Taichi. It is a bit clear to see that Taichi actions. \\ \hline
2 & No. No. No.  & A bit. A bit. No \\ \hline
3 & 1. No 2. Jumping jacks 3. disorted face & 1. litle 2. no 3. disorted hands \\ \hline
4 & no, no, no & almost (general shape is recognizable), action is something like yoga \\ \hline
5 & no, no & smth between, no \\ \hline
6 & no & yes \\ \hline
7 & No, no, no & No, no, no \\ \hline
8 & The body is okay, it only gets weird for a bit. I’m not sure what the person is doing. & The bods is okay and the action looks like some slow motion of mopping the floor. \\ \hline
9 & 1. No, 2. No & 1. No, 2. No \\ \hline
10 & Not photorealistic / no & Not photorealistic / no \\ \hline
11 & I understand what it’s supposed to do. However, I’m not sure if it’s working clearly because there’s no movement in the legs. & His movements are so small and slow that I can’t tell what he’s doing. It looks like Tai Chi, but if I didn’t know it was supposed to be Tai Chi, I wouldn’t have guessed. \\ \hline
12 & body is rendered, but it's scuffed. Identity is consistent. Action is identifiable but not clearly. & Guy is not photorealistic. It is consistent. His action is not identifiable. \\ \hline
13 & The person is not photorealistic. Action seems to be jumping jack but might be something else. & Person is quite photorealistic, but action hard to recognize. \\ \hline
14 & not photorealistic, dont know what he is doing, & body rendered properly, action not ideantifiable \\ \hline
15 & No, Yes, Why & No, No, Why \\ \hline
16 & 1. Not photorealistic 2. Is quite identifiable & 1. Not photorealistic 2. Not identifiable \\ \hline
\end{tabular}
\caption{Free-form participant responses to questions about the quality of two videos generated by \textit{AnimateAnyone} (Q3.1 and Q3.2 within Task 3: Qualitative Evaluation).}
\label{tab:qual-results-1}
\end{table*}

\begin{table*}[h]
\centering
\begin{tabular}{|l|p{0.4\textwidth}|p{0.4\textwidth}|}
\hline
\textbf{ID} & \textbf{Q3.3} & \textbf{Q3.4} \\ \hline
1 & There is an old lady performing body weight squats. She is photorealistic. She looks like a very old woman, around 80 - 100 years old. & There is a women doing body weight squats. The AI sometimes cant animate her foot and head properly. The video is fairly photorealistic. \\ \hline
2 & No. yes. No & No. A bit. No \\ \hline
3 & 1. yes 2. yes 3. Two people & 1. partitialy 2. yes 3. no \\ \hline
4 & This is the better one, action is recognizable  & No, action is recognizable  \\ \hline
5 & yes, yes & yes, yes \\ \hline
6 & yes & no \\ \hline
7 & No, yes, no & No, no, no \\ \hline
8 & The action is clear here, but when the person squats another person appears behind them and it’s weird. & The hair changes and she’s missing one foot. But the action is clear. \\ \hline
9 & 1. Yes, 2. Yes & 1. No, 2. Yes \\ \hline
10 & Not photorealistic / yes & Not photorealistic / yes \\ \hline
11 & It’s very clear that she’s doing squats. There’s a bit of an afterimage, but the movement is still obvious. & It’s clear that she’s doing squats. The hand movements aren’t important in this action. \\ \hline
12 & This is not photorealistic but better than most of the others. It has consistency. Action is clearly identifiable. & That is not photorealistic. It not very consistent. Action is clearly identifiable. \\ \hline
13 & Person quite photorealistic. Action recognizable, but the figure of person behind them makes it puzzling. & Person photorealistic only in certain segments of the video, action vaguely recognizable. \\ \hline
14 & body almost rendered properly, identifiable action, would be perfect if not the silhuette in background & body not photorealistic, action somewhat identifiable \\ \hline
15 & No, Yes, Why & No, Yes, Why \\ \hline
16 & 1. Not photorealistic 2. identifiable 3. Person is duplicated mid action & 1. Quite photorealistic clothes, but distorted face 2. Identifiable \\ \hline
\end{tabular}
\caption{Free-form participant responses to questions about the quality of two videos generated by \textit{MagicAnimate} (Q3.3 and Q3.4 within Task 3: Qualitative Evaluation).}
\label{tab:qual-results-2}
\end{table*}

\begin{table*}[h]
\centering
\begin{tabular}{|l|p{0.4\textwidth}|p{0.4\textwidth}|}
\hline
\textbf{ID} & \textbf{Q3.5} & \textbf{Q3.6} \\ \hline
1 & A man is clearly performing Taichi, he is photorealistic and the render is not that good in this video. & A man doing table tennis, the animation of his actions are not that good, but after a closer look, it was obvious. He is photorealistic and the render is not good once again in this video. \\ \hline
2 & Yes. Yes. No & Yes. Yes. No \\ \hline
3 & 1. yes 2. no 3. no & 1. yes 2. yes 3. no \\ \hline
4 & No, action is better & no but action is also better \\ \hline
5 & yes, no & yes, yes \\ \hline
6 & yes & yes \\ \hline
7 & Yes, no, no & Yes, no, no \\ \hline
8 & The body is good but I have no clue what the action is. & The body and action are both good. \\ \hline
9 & 1. No, 2. Yes & 1. No, 2. Yes \\ \hline
10 & photorealistic / yes & Photorealistic / yes \\ \hline
11 & Hmm… Since there’s no background, the movements are easy to see, but they’re so slow and small that it’s not clear what he’s doing. & Hmm… Since there’s no background, the movements are easy to see, but they’re so slow and small that it’s not clear what he’s doing. \\ \hline
12 & It is not photorealistic. It is consistent. Action is not identifiable. & Person is not photorealistic. It is consistent. It is not clear what he is doing.  \\ \hline
13 & Person not very recognizable, action hard to recognize. & Person not photorealistic, action very vague and hard to identify. \\ \hline
14 & is photorealistic, action probably identifiable, if i didnt remebmer that taichi video, i wouldnt know & photorealistic, identifiable action,  \\ \hline
15 & No, No, Why & No, Yes, Why \\ \hline
16 & 1. Not photorealistic 2. Not identifiable & 1. Not photorealistic, but quite good quality 2. Identifiable \\ \hline
\end{tabular}
\caption{Free-form participant responses to questions about the quality of two videos generated by \textit{ExAvatar} (Q3.5 and Q3.6 within Task 3: Qualitative Evaluation).}
\label{tab:qual-results-3}
\end{table*}

\end{document}